\documentclass[sn-mathphys-num]{sn-jnl}

\usepackage{graphicx}%
\usepackage{multirow}%
\usepackage{amsmath,amssymb,amsfonts}%
\usepackage{amsthm}%
\usepackage{mathrsfs}%
\usepackage[title]{appendix}%
\usepackage{textcomp}%
\usepackage{manyfoot}%
\usepackage{booktabs}%
\usepackage{algorithm}%
\usepackage{algorithmicx}%
\usepackage{algpseudocode}%
\usepackage{listings}%
\usepackage{caption}%
\usepackage{silence}
\usepackage[table]{xcolor}

\begin{document}

\title[Article Title]{PointOBB-v3: Expanding Performance Boundaries of Single Point-Supervised Oriented Object Detection}


\author[1]{\fnm{Peiyuan} \sur{Zhang}} \email{peiyuanzhangwhu@whu.edu.cn}
\equalcont{These authors contributed equally to this work.}

\author[2]{\fnm{Junwei} \sur{Luo}} \email{luojunwei@whu.edu.cn}
\equalcont{These authors contributed equally to this work.}

\author[3]{\fnm{Xue} \sur{Yang}} \email{yangxue-2019-sjtu@sjtu.edu.cn}
\equalcont{These authors contributed equally to this work.}

\author[4]{\fnm{Yi} \sur{Yu}} 

\author[5]{\fnm{Qingyun} \sur{Li}}

\author[6]{\fnm{Yue} \sur{Zhou}}

\author[7]{\fnm{Xiaosong} \sur{Jia}}

\author[8]{\fnm{Xudong} \sur{Lu}}

\author[9]{\fnm{Jingdong} \sur{Chen}}

\author[10]{\fnm{Xiang} \sur{Li}}

\author[7]{\fnm{Junchi} \sur{Yan}}

\author*[2]{\fnm{Yansheng} \sur{Li}}\email{yansheng.li@whu.edu.cn}

\affil[1]{\normalsize \orgdiv{School of Computer Science}, \orgname{Wuhan University}, \orgaddress{\city{Wuhan}, \country{China}}}

\affil[2]{\normalsize \orgdiv{School of Remote Sensing and Information Engineering}, \orgname{Wuhan University}, \orgaddress{\city{Wuhan}, \country{China}}}

\affil[3]{\normalsize \orgdiv{Department of Automation}, \orgname{Shanghai Jiao Tong University}, \orgaddress{\city{Shanghai}, \country{China}}}

\affil[4]{\normalsize \orgdiv{School of Automation}, \orgname{Southeast University}, \orgaddress{\city{Nanjing}, \country{China}}}

\affil[5]{\normalsize \orgdiv{SEIE}, \orgname{Harbin Institute of Technology}, \orgaddress{\city{Harbin}, \country{China}}}

\affil[6]{\normalsize \orgdiv{S-Lab, CCDS}, \orgname{Nanyang Technological University}, \orgaddress{\country{Singapore}}}

\affil[7]{\normalsize \orgdiv{CSE \& SAI}, \orgname{Shanghai Jiao Tong University}, \orgaddress{\city{Shanghai}, \country{China}}}

\affil[8]{\normalsize \orgname{The Chinese University of Hong Kong}, \orgaddress{\city{Hong Kong SAR}, \country{China}}}

\affil[9]{\normalsize \orgname{Ant Group}, \orgaddress{\city{Hangzhou}, \country{China}}}

\affil[10]{\normalsize \orgdiv{VCIP, CS}, \orgname{Nankai University}, \orgaddress{\city{Tianjin}, \country{China}}}

\abstract{With the growing demand for oriented object detection (OOD), recent studies on point-supervised OOD have attracted significant interest. In this paper, we propose PointOBB-v3, a stronger single point-supervised OOD framework. Compared to existing methods, it generates pseudo rotated boxes without additional priors and incorporates support for the end-to-end paradigm. PointOBB-v3 functions by integrating three unique image views: the original view, a resized view, and a rotated/flipped (rot/flp) view. Based on the views, a scale augmentation module and an angle acquisition module are constructed. In the first module, a Scale-Sensitive Consistency (SSC) loss and a Scale-Sensitive Feature Fusion (SSFF) module are introduced to improve the model's ability to estimate object scale. To achieve precise angle predictions, the second module employs symmetry-based self-supervised learning. Additionally, we introduce an end-to-end version that eliminates the pseudo-label generation process by integrating a detector branch and introduces an Instance-Aware Weighting (IAW) strategy to focus on high-quality predictions. We conducted extensive experiments on the DIOR-R, DOTA-v1.0/v1.5/v2.0, FAIR1M, STAR, and RSAR datasets. Across all these datasets, our method achieves an average improvement in accuracy of 3.56\% in comparison to previous state-of-the-art methods. The code will be available at \href{https://github.com/ZpyWHU/PointOBB-v3.git}{https://github.com/ZpyWHU/PointOBB-v3}.}
\keywords{Oriented object detection, weakly-supervised learning, remote sensing}

\maketitle

\section{Introduction}\label{sec1}
Oriented object detection (OOD) in aerial imagery focuses on identifying and localizing objects of interest by employing oriented bounding boxes (OBBs), while also determining their respective categories. This area has seen a significant amount of high-quality research contributions \cite{Ding_Xue_Long_Xia_Lu_2019,Han_Ding_Xue_Xia_2021,Li_Hou_Zheng_Cheng_Yang_Li_2023,Xie_Cheng_Wang_Yao_Han_2021,Yang_Yan_Feng_He_2022,li2023learningholisticallydetectbridges,Yang_Yan_Qi_Wang_Zhang_Tian_2021,Yang_Yang_Yang_Qi_Wang_Tian_Yan_2021,li2025simple}. Despite these advancements, the process of manually annotating detailed OBBs remains labor-intensive and costly, posing challenges for large-scale datasets. 

\begin{figure}[h]
    \centering
    \includegraphics[width=1\textwidth]{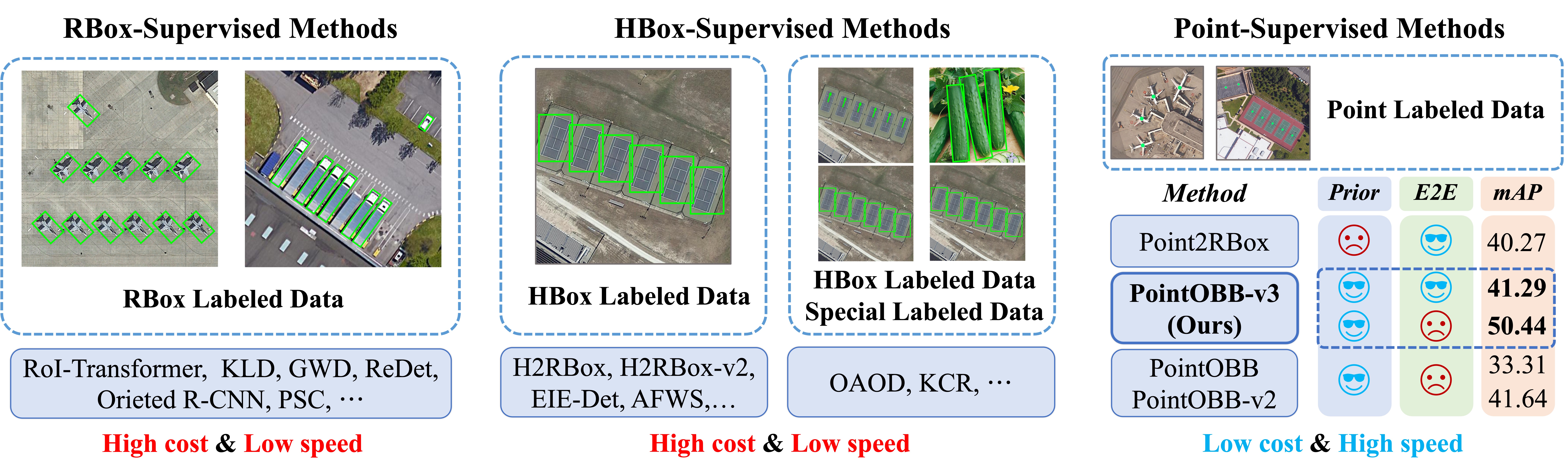}
    \caption{The main paradigmatic types of existing OOD. Compared to RBox and HBox labels, point labels have lower costs and higher efficiency. ``Prior'' indicates using additional human knowledge priors. ``E2E'' indicates training in an end-to-end manner, and ``mAP'' shows performance on the DOTA-v1.0 dataset using mAP$_{50}$ metric.}
    \label{fig:intro}
\end{figure}

To reduce the annotation cost of bounding boxes, weakly-supervised horizontal object detection utilizing image-level annotations has been extensively developed \cite{Bilen_Vedaldi_2016,Chen_Fu_Jiang_Chen_Hua_2020,Feng_Yao_Cheng_Han,Feng_Yao_Shen_Han_Feng_Cheng_Xiao,Tang_Wang_Bai_Liu_2017,Tang_Wang_Bai_Shen_Bai_Liu_Yuille_2020,Wan_Wei_Han_Jiao_Ye_2019}. However, these approaches struggle in complex aerial scenes and lack the capability to predict object orientations. In recent years, there has been increasing interest in weakly supervised OOD. As depicted in Fig.~\ref{fig:intro}, existing weakly-supervised methods use coarser-grained annotations as weak supervisory signals to predict OBBs, such as horizontal bounding box (HBB) annotations \cite{Sun_Ran_Yang_Gao_Kurozumi_Kimata_Ye_2021,Yang_Zhang_Li_Wang_Zhou_Yan_2022,Yu_Yang_Li_Zhou_Zhang_Yan_Da_2023, Zhu_Ferenczi_Purkait_Drummond_Rezatofighi_Hengel_2023}. Nevertheless, box-based annotations remain inefficient and labor-intensive. Thus, exploring more cost-effective and efficient annotation forms is crucial.

Recently, point-based annotation has garnered considerable attention in various tasks \cite{Cheng_Parkhi_Kirillov_2022,Fan_Zhang_Tan_2022,Li_Yuan_Wang_Zhu_Li_Liu_Zhang,Ren_Yu_Yang_Liu_Schwing_Kautz_2020}. Within the realm of object detection, point annotations have proven to be approximately 36.5\% less costly compared to HBBs and 104.8\% less costly compared to OBBs, while increasing the labeling efficiency compared to both HBB and OBB. As a result, single point-supervised OOD emerges as a more meaningful and efficient alternative. 

From a bird's-eye view, objects in aerial images exhibit two distinct characteristics: varying spatial scales and arbitrary orientations. Given the abundance of small objects in aerial imagery \cite{Xia_Bai_Ding_Zhu_Belongie_Luo_Datcu_Pelillo_Zhang_2018}, utilizing single point labels is clearly more suitable than using multiple point labels as in methods like \cite{Cheng_Parkhi_Kirillov_2022}. Current single point-supervised object detection techniques \cite{Chen_Yu_Han_Hassan_Wang_Li_Zhao_Shi_Han_Ye_2022,Papadopoulos_Uijlings_Keller_Ferrari_2017} follow the Multiple Instance Learning (MIL) paradigm. This approach optimizes based on category labels, selecting the proposals with the highest confidence from proposal sets as the predicted bounding boxes, thereby achieving perception of object scale. However, the MIL paradigm inherently struggles with stability in perceiving object boundaries, often focusing on the most distinctive part of an object rather than its full extent and precise boundaries \cite{Tang_Wang_Bai_Liu_2017,Tang_Wang_Bai_Shen_Bai_Liu_Yuille_2020}. We identify two key challenges when extending this approach to OOD in aerial imagery: i) How to address the inconsistency in the MIL approach to obtain more accurate scale representations? ii) How to effectively learn the object's orientation under single point-supervision?

In this paper, we introduce a stronger single \textbf{Point}-based \textbf{OBB} generation framework termed \textbf{PointOBB-v3}. It facilitates collaborative learning of both angle and scale by integrating three distinct views. The main design consists of the following five key aspects: a progressive multi-view switching strategy, Scale-Sensitive Consistency (SSC) loss, Scale-Sensitive Feature Fusion (SSFF) module, Self-Supervised Angle (SSA) loss and Dense-to-Sparse (DS) matching strategy. The progressive multi-view switching strategy aims to enable the model to progressively learn the object's orientation and scale across three views. SSC loss is used to address the aforementioned inconsistency between the proposal's confidence score and its scale accuracy. The SSFF module is designed to further enhance scale learning. The purpose of the SSA loss and DS matching strategy is to obtain accurate object angle prediction.

Specifically, to collaboratively learn both the angle and scale of objects, we construct three views from the input image and point label: the original view, and two enhanced views—one resized and one rotated/flipped (rot/flp). Meanwhile, to facilitate the progressive acquisition of the network's discriminative capabilities for object scale and orientation, we implement a progressive multi-view switching strategy during training, wherein the two enhanced views are alternated properly to allow the model to gradually learn both scale and orientation knowledge.

To tackle the two aforementioned issues, we propose two modules based on the three views: a scale enhancement module using the original and resized views, and an angle acquisition module using the original and rot/flp views. i) \textbf{The scale enhancement module.} The core components of this module include the SSC loss and the SSFF module. In an optimal scenario, the predicted size of the same object---reflected by the scale of the proposal with the highest confidence score---should remain consistent across views with different resolutions. To enforce this consistency, the SSC loss is designed to reduce the divergence in the predicted score distributions between the original and resized views. Furthermore, feature layer mapping in traditional MIL-based approaches is strictly based on ROI’s scale and predefined hyperparameters, which can be inflexible and unstable, especially when the scale is imprecise. This often leads to incorrect layer mapping and inconsistent feature extraction. To address this, we propose the SSFF module that uses a gating mechanism inspired by Mixture of Experts (MoE) architecture. This mechanism generates gating scores for each feature layer, which are then used to dynamically aggregate the multiple output layers of the FPN, enhancing the model's ability to perceive scale information. ii) \textbf{The angle acquisition module.} Within this module, the SSA loss is designed based on the inherent symmetry of objects in aerial images. To obtain more accurate OBBs, we propose the DS matching strategy that complements the self-supervised angle learning branch in this module.

Additionally, existing point-supervised methods~\cite{luo2024pointobb, ren2024pointobbv2simplerfasterstronger} commonly adopt a two-stage paradigm, where pseudo-labels are first generated and then used for detector training. Considering the complexity and extended training time associated with the two-stage paradigm, we further implement an end-to-end version of our proposed PointOBB-v3 framework. This end-to-end version primarily integrates a Rotated FCOS~\cite{Tian_Shen_Chen_He_2019}-based detection head, forming a dual-branch structure with the MIL head to enable joint training. Moreover, an Instance-Aware Weighting (IAW) strategy is proposed to alleviate the negative impact of low-quality predictions from the MIL head on the training process. This strategy dynamically applies instance-level weight to the end-to-end loss, effectively improving parameter optimization during joint training.

This paper is an extended version of our conference paper, published in CVPR 2024 \cite{luo2024pointobb}, and the main contributions are as follows \footnote{This paper is built upon our work published in CVPR 2024 \cite{luo2024pointobb} with substantial extensions. Compared to the original version, we extend PointOBB in five aspects. \textbf{1)} We propose an SSFF module to further enhance the model's capability to perceive object scales (in Sec. \ref{subsec6}). Experiments demonstrate that the proposed SSFF module significantly improves the performance of PointOBB, outperforming existing single point-supervised methods. For instance, PointOBB-v3 achieves a 2.20\% improvement over PointOBB-v2 \cite{ren2024pointobbv2simplerfasterstronger} on DIOR-R and an 8.76\% improvement over PointOBB-v2 on DOTA-v1.0. \textbf{2)} We also propose an end-to-end version of the framework that directly outputs prediction results without the need for generating pseudo-labels (in Sec. \ref{subsec8}). Comprehensive experiments have been conducted to validate the effectiveness of this end-to-end framework. For example, it achieves a performance of 41.29\% on the DOTA-v1.0 dataset, even surpassing two-stage methods like PointOBB. \textbf{3)} We conducted additional experiments on a wider range of datasets, including DIOR, DOTA-v1.0, DOTA-v1.5, DOTA-v2.0, FAIR1M, STAR, and RSAR. Across all these datasets, our method achieves an average improvement in accuracy of 3.56\% in comparison to previous state-of-the-art methods (in Sec. \ref{subsubsec12}). \textbf{4)} We provide more ablation studies to further understand the design principles and capability of PointOBB-v3. (in Sec. \ref{subsec12}) \textbf{5)} We have refined the manuscript by incorporating more intuitive figures and reorganizing the content and tables to enhance clarity and readability.}: \textbf{1)} The proposed approach integrates object scale and orientation learning across three unique perspectives, guided by a progressive multi-view switching strategy. \textbf{2)} An SSC loss is devised to strengthen the network's capability in capturing object scale, and an SSFF module is designed to enhance accuracy in scale-related feature layer mapping. \textbf{3)} An SSA loss is used to learn the object orientation and a scale-guided DS matching strategy is proposed to further enhance the precision of object angle prediction. \textbf{4)} Based on the aforementioned principles, we implemented two frameworks: one end-to-end and the other two-stage, providing researchers and practitioners with a wider range of options for further exploration and application. \textbf{5)} We have made improvements in the manuscript, including conducting experiments on additional benchmarks, providing deeper insights into performance improvements, conducting more comprehensive ablation studies, and reorganizing the contents and tables for better clarity.

\section{Related work}\label{sec2}
\subsection{Fully-Supervised Oriented Object Detection}\label{subsec1}
OOD algorithms primarily target different types of objects such as aerial objects \cite{Xu_Fu_Wang_Wang_Chen_Xia_Bai_2021}, multi-oriented scene texts \cite{Liao_Zhu_Shi_Xia_Bai_2018} \cite{Liu_Liang_Yan_Chen_Qiao_Yan_2018}, and 3D objects \cite{Yang_Zhang_Yang_Zhou_Wang_Tang_He_Yan_2022}. Among the representative approaches are anchor-free detectors like Rotated FCOS \cite{Tian_Shen_Chen_He_2019}, and anchor-based detectors including Rotated RetinaNet \cite{Lin_Goyal_Girshick_He_Dollar_2017}, RoI Transformer \cite{Ding_Xue_Long_Xia_Lu_2019}, Oriented R-CNN\cite{Xie_Cheng_Wang_Yao_Han_2021}, and ReDet \cite{Han_Ding_Xue_Xia_2021}.
Oriented RepPoints \cite{Li_Chen_Hu_Zhu} presents a strategy for evaluating and allocating sample quality based on adaptive points. Additionally, methods such as R$^{3}$Det \cite{Yang_Yan_Feng_He_2022} and S$^{2}$A-Net \cite{Han_Ding_Li_Xia_2022} boost detector performance by utilizing feature alignment modules. To overcome boundary discontinuity issues during angle regression, methods using angle coders \cite{Yang_Yan_2020, Yang_Hou_Zhou_Wang_Yan_2021, Yu_Da_2022} transform angles into boundary-free formats, thus enhancing stability. Moreover, approaches like GWD \cite{Yang_Yan_Qi_Wang_Zhang_Tian_2021} and KLD \cite{Yang_Yang_Yang_Qi_Wang_Tian_Yan_2021} propose Gaussian-based loss functions to effectively analyze the nature of rotational representations, thereby improving overall detection performance.

\subsection{Weakly-Supervised Oriented Object Detection}\label{subsec2}
Existing weakly-supervised OOD approaches can be broadly categorized into image-supervised and HBox-supervised methods. Additionally, we further explore the feasibility of point-supervised methods as a more efficient alternative, aiming to reduce annotation costs while maintaining high detection performance.

\textbf{Image-supervised.}
For methods using image-level annotation, WSODet \cite{Tan_Jiang_Guo_Zhang_2023} enhances the OICR \cite{Tang_Wang_Bai_Liu_2017} framework to predict HBBs. Subsequently, pseudo OBBs are generated using contour features and the predicted HBBs, followed by an Oriented RepPoints branch to refine these predictions. However, image-supervised methods generally yield limited performance, as the quality of the generated OBBs heavily relies on the accuracy of the predicted HBBs, which can lead to suboptimal results.

\textbf{HBox-supervised.}
For methods using HBox-level annotation, although some approaches (e.g., BoxInst-RBox \cite{Tian_Shen_Wang_Chen_2021} and BoxLevelSet-RBox \cite{Li_Liu_Zhu_Cui_Hua_Zhang}) follow an HBox-Mask-RBox pipeline to generate OBBs, involving segmentation often results in higher computational costs, making the entire procedure time-consuming. H2RBox \cite{Yang_Zhang_Li_Wang_Zhou_Yan_2022} takes a more efficient approach by predicting RBoxes directly from HBox annotations without relying on redundant representations. This method learns the angle from the geometry of circumscribed boxes, achieving notable performance. H2RBox-v2 \cite{Yu_Yang_Li_Zhou_Zhang_Yan_Da_2023} further improves upon this by leveraging the inherent symmetry of objects. However, these methods still require collecting a substantial number of bounding box annotations. In addition to these, some studies leverage HBox along with specialized forms of annotations. OAOD \cite{iqbal2021leveragingorientationweaklysupervised} utilizes additional object angle annotations, while KCR \cite{Zhu_Ferenczi_Purkait_Drummond_Rezatofighi_Hengel_2023} employs a combination of RBox-annotated source datasets and HBox-annotated target datasets. Sun et al. \cite{Sun_Ran_Yang_Gao_Kurozumi_Kimata_Ye_2021} integrate HBox annotations with image rotations to align oriented objects horizontally or vertically. Nevertheless, such specialized forms of annotation lack universality, limiting their applicability to a broader range of tasks.

\textbf{Point-supervised.}
Point-based annotations have been widely used in various tasks such as object detection \cite{Chen_Yang_Zhang_Zhang_Sun_2021, Chen_Yu_Han_Hassan_Wang_Li_Zhao_Shi_Han_Ye_2022, He_Zou_Wang_Li_Cao_Jing, Papadopoulos_Uijlings_Keller_Ferrari_2017, Ren_Yu_Yang_Liu_Schwing_Kautz_2020, Ying_Liu_Wang_Li_Chen_Lin_Sheng_Zhou_2023, yu2025point2rbox}, panoptic segmentation \cite{Fan_Zhang_Tan_2022, Li_Yuan_Wang_Zhu_Li_Liu_Zhang}, and instance segmentation \cite{Bearman_Russakovsky_Ferrari_Fei-Fei_2016, Cheng_Parkhi_Kirillov_2022}, among others \cite{iqbal2021leveragingorientationweaklysupervised, Yu_Chen_Wu_Hassan_Li_Yan_Shi_Ye_Han}. Due to its cost-effectiveness and efficiency, single point-supervised object detection has garnered significant attention.  Click \cite{Papadopoulos_Uijlings_Keller_Ferrari_2017} was one of the earliest attempts at point-supervised object detection, proposing center-click annotations and incorporating them into an MIL paradigm. P2BNet \cite{Chen_Yu_Han_Hassan_Wang_Li_Zhao_Shi_Han_Ye_2022} further enhances this approach by using a coarse-to-fine strategy and incorporating negative samples to improve prediction quality. However, these methods only produce horizontal bounding boxes and fail to address the inherent instability of the MIL paradigm effectively. 

Based on the above, to ultimately obtain RBox from point annotations, one potential method is the Point-to-Mask approach \cite{Li_Yuan_Wang_Zhu_Li_Liu_Zhang}, which involves determining the minimum circumscribed rectangle of the generated mask. Another viable method entails a combination of Point-to-HBox and HBox-to-RBox strategies. However, in the past year, several innovative methods have emerged, offering new insights into point-supervised OOD. Point2RBox \cite{yu2024point2rbox} introduces an end-to-end approach that combines knowledge and learns from one-shot examples to generate RBox. And PointOBB-v2 \cite{ren2024pointobbv2simplerfasterstronger} learns a class probability map to generate pseudo RBox labels through principal component analysis. In our experiments, we employ these novel approaches for comparison. Overall, we propose a stronger framework for single point-supervised OOD based on the MIL paradigm, providing an efficient approach to directly achieve oriented object detection via single point-supervision.

\section{Preliminaries}\label{sec3}
\subsection{MIL-Based Weakly-Supervised Object Detection}\label{subsec3}
As a classic weakly supervised learning method, MIL paradigm \cite{dietterich1997solving,zhou2004multi} has been widely applied in various weakly supervised tasks like image-supervised \cite{bilen2016weakly,tang2017multiple,tang2018pcl,chen2020slv,shen2020uwsod,zhang2021weakly,feng2022weakly,feng2023learning} and point-supervised horizontal object detection \cite{papadopoulos2017training,ren2020ufo,Chen_Yu_Han_Hassan_Wang_Li_Zhao_Shi_Han_Ye_2022}. 

The fundamental design of the MIL paradigm is the two-stream architecture, which predicts a set of aggregate scores $S$ consisting of the instance scores and the class scores, from a set of candidate proposal bags. The proposal bag with the highest aggregate score is selected under the constraint of classification labels, serving as the prediction for the object's category and scale. In the single point-supervised object detection setting, each object has a point label that includes the spatial coordinate and the category, and a candidate proposal bag is constructed based on the point label. Generally, the common process of MIL under such a setting can be formalized as:

\begin{equation}
\mathcal{L}_{\text{mil}} = \sum_{j=1}^{J} \sum_{k=1}^{K} \big( [c_j]_k \log([{S}_j]_k) + (1 - [c_j]_k) \log(1 - [{S}_j]_k) \big),
\label{eq:mil1_loss}
\end{equation}
where \textit{J} denotes the number of proposal bags in the batch, \textit{K} represents the total number of categories, $c_j \in \{0, 1\}^K$ is the one-hot category label, and $[{S}_j]_k$ refers to the aggregate score for the \textit{k}-th category of the \textit{j}-th proposal in the ${S}$. 

Above all, by optimizing the aforementioned MIL-based paradigms, the network acquires critical perceptual capabilities for object scales and serves as the foundational architecture for our method.

\subsection{Symmetry-Aware Learning}\label{subsec4}
Symmetry is a natural attribute widely present in various scenarios. Therefore, learning the orientation of objects through their symmetry is theoretically feasible in weakly supervised object detection \cite{Yu_Yang_Li_Zhou_Zhang_Yan_Da_2023}. Assume there is a neural network \textit{f}$_{nn}$(·) that maps a symmetric image \textit{I} to a real number \textit{$\theta$}, $\theta = f_{nn}(I)$, To leverage reflection symmetry, the function is enhanced with two additional properties: flip consistency and rotate consistency. 

\textbf{i. Flip consistency}. When the input image is flipped vertically, \textit{f}$_{nn}$(·) gives an opposite output:
\begin{equation}
f_{nn}(I) + f_{nn}(\text{flp}(I)) = 0,
\label{eq:flip consistency}
\end{equation}
where \text{flp}(\textit{I}) is an operator of vertically flipping the image \textit{I}.

\textbf{ii. Rotate consistency}. When the input image is rotated by $\mathcal{R}$,  the output of \textit{f}$_{nn}$(·) also rotates by $\mathcal{R}$:
\begin{equation}
f_{nn}(\text{rot}(I, \mathcal{R})) - f_{nn}(I) = \mathcal{R},
\label{eq:rotate consistency}
\end{equation}
where \text{rot}(\textit{I}, $\mathcal{R}$) is an operator that clockwise rotates the image \textit{I} by $\mathcal{R}$. Given an image \textit{I$_0$} symmetric about $\varphi = \theta_{\text{sym}}$, assuming the corresponding output is $\theta_{\text{pred}} = f_{nn}(I_0)$, the image can be transformed in two ways: i) Flipping along the line $\varphi = \theta_{\text{sym}}$. ii) Flipping vertically first, and then rotating by 2$\theta_{\text{sym}}$. Both transformations result in the output of \textit{f}$_{nn}$(·) remaining the same, which leads to \( \theta_{\text{pred}} = \theta_{\text{sym}} \). Therefore, when the network successfully learns these consistency properties, its output will align precisely with the orientation of the image's axis of symmetry. As described in Sec. \ref{subsec7}, we employ symmetry and design an assigner to pair objects across different views, which facilitates the computation of consistency loss for these matched objects and allows for generalization to multiple object detection tasks.

\begin{figure}[!tb]
    \centering
    \includegraphics[width=1\textwidth]{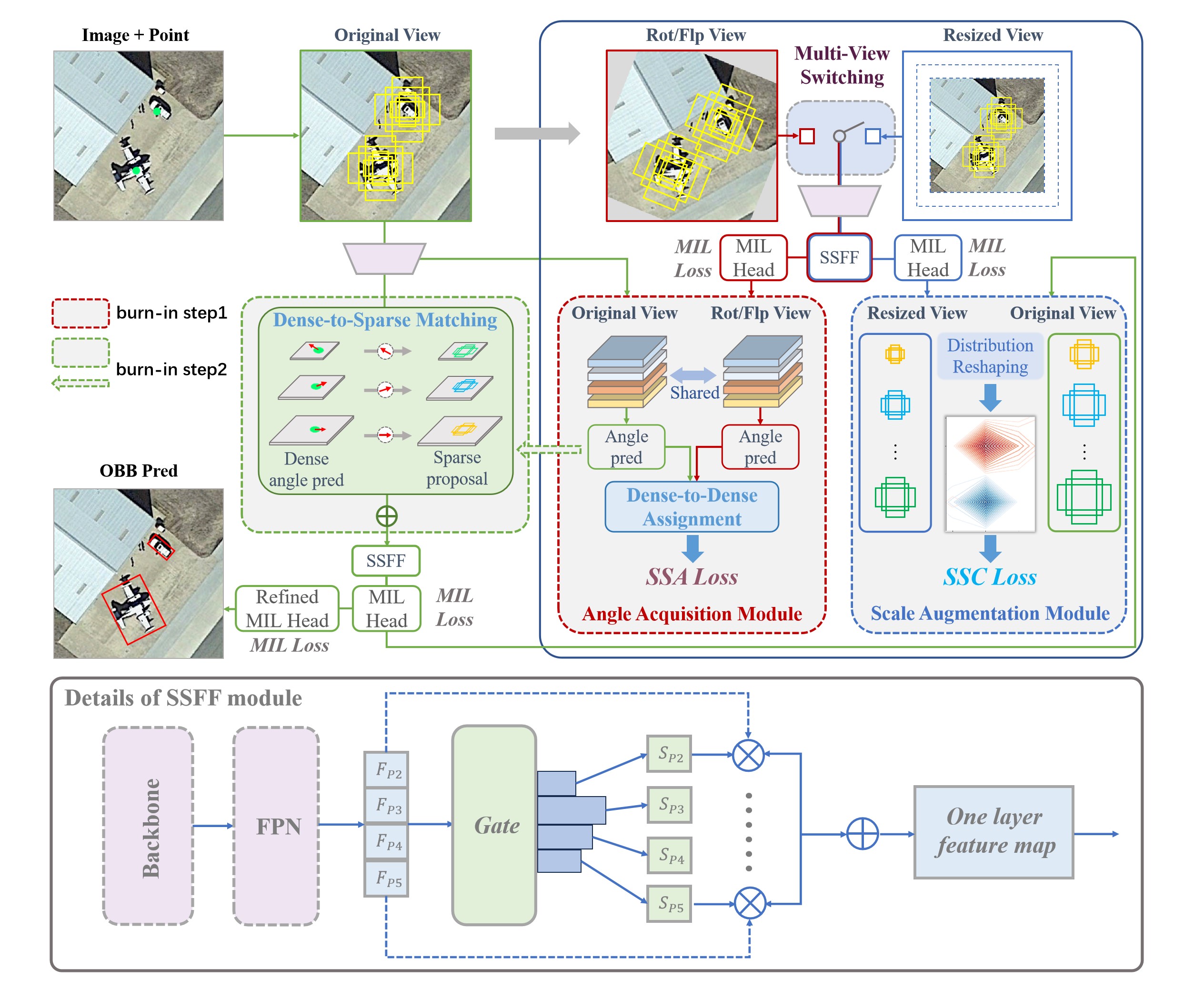}
    \caption{The pipeline of the PointOBB-v3. PointOBB-v3 consists of three distinct views, which serve as the foundation for constructing an angle acquisition module and a scale augmentation module. The latter integrates a Scale-Sensitive Consistency (SSC) loss and a Scale-Sensitive Feature Fusion (SSFF) module to improve the model's capability in perceiving scale variations. The angle acquisition module incorporates a Dense-to-Dense sample assignment mechanism to facilitate angle learning, optimized through the Self-Supervised Angle (SSA) loss. Moreover, a Dense-to-Sparse (DS) matching strategy is introduced to achieve more accurate object angle estimation. During training, a progressive multi-view switching strategy is implemented, enabling the transition between resized views, rotated/flipped views, and their associated modules.}
    \label{fig:main_method}
\end{figure}

\section{Methodology}\label{sec4}

\subsection{Overall Framework}\label{subsec5} 
This section provides a comprehensive overview of the proposed PointOBB-v3. Sec. \ref{subsec6} delves into the scale augmentation module, including an in-depth explanation of the SSC loss and the SSFF module. In Sec. \ref{subsec7}, we explore the angle acquisition module in detail. Sec. \ref{subsec8} focuses on the design and implementation of the end-to-end framework, while Sec. \ref{subsec9} concludes with a discussion of the overall optimization objective with the MIL loss.

\subsubsection{Method Overview}\label{subsubsec1}
In current research on image-supervised and point-supervised object detection, MIL-based paradigms demonstrate fundamental perceptual capabilities for object scale. Building upon this foundation, we adopt the classic MIL fashion as the underlying network for our approach. The overall framework of PointOBB-v3 is illustrated in Fig.~\ref{fig:main_method}. From the original view, the generation of initial proposal bags is guided by the provided point labels. Subsequently, angle predictions obtained from the angle acquisition module are matched to these proposals through the DS matching strategy, thereby endowing the horizontal proposals with orientation information. From the generated rotated proposals, a subset of reliable proposals is selected and progressively refined using a MIL head followed by a refined MIL head, ultimately yielding the final OBB predictions. The obtained OBBs can be utilized as pseudo-labels to facilitate the final training stage of oriented object detectors.

The above pipeline incorporates three distinct views. Based on the original view, a resized view is generated by applying random scaling, and a rot/flp view is derived by performing random rotations or vertical flipping. These three views are leveraged to design two critical modules: the scale augmentation module and the angle acquisition module. The scale augmentation module is tailored to improve the network’s ability to perceive and adapt to variations in object scales, whereas the angle acquisition module focuses on effectively learning object orientations through symmetry-aware learning. The resized and rot/flp views act as augmented views, which are dynamically alternated during training using the proposed progressive multi-view switching strategy ensuring a more robust and adaptive learning process.

Besides, while the two-stage training paradigm with pseudo-label generation demonstrates strong performance, it incurs significant computational cost and extended training time. To address this, we also developed an end-to-end framework based on the same principles. This end-to-end version achieves a substantial improvement in training efficiency, reducing training hours by 21.36\%, while maintaining competitive detection performance. The proposed point-supervised rotation object detection frameworks offer distinct advantages: the two-stage approach provides higher precision, while the end-to-end framework emphasizes efficiency and convenience. Together, these options aim to provide researchers and practitioners with a flexible and diverse set of tools to cater to various application requirements.

\subsubsection{Progressive Multi-View Switching Strategy}\label{subsubsec2}
To facilitate the network's gradual development of discriminative capabilities for object scale and predictive abilities for object orientation, we introduce a progressive multi-view switching strategy as a core optimization mechanism in our framework. 

This progressive multi-view switching strategy operates through three stages: \textbf{Stage 1}. The process begins by generating a resized view from the original view using a scaling factor $\sigma$. This resized view, combined with the original view, adheres to scale-equivalence constraints, forming the basis of the scale augmentation module. This module is designed to improve the network’s capability to accurately perceive object scales, laying the groundwork for effective scale-sensitive learning. Details of this module will be discussed in subsequent sections. \textbf{Stage 2} (i.e., ``burn-in step1" in Fig.~\ref{fig:main_method}). Before this stage, the network has developed fundamental perceptual abilities to the scale and boundary of objects. However, orientation information is still lacking. To address this, the resized view is replaced with a rot/flp view, which is paired with the original view to construct the angle acquisition module. This module employs a dense-to-dense sample assignment approach, enabling the network to engage in self-supervised angle learning, thereby introducing orientation sensitivity. \textbf{Stage 3} (i.e., ``burn-in step2" in Fig.~\ref{fig:main_method}). At this stage, the network has achieved a refined ability to predict accurate angles using dense feature representations. To integrate these angle predictions with object proposals, the proposed DS matching strategy is employed. This strategy aligns dense angle predictions with sparse proposals by leveraging neighboring receptive fields, effectively assigning accurate object orientations to the proposals.

\subsection{Scale Augmentation Module}\label{subsec6}
\subsubsection{Scale-Sensitive Consistency Loss}\label{subsubsec3}
Objects in aerial images often display substantial scale variations, posing significant challenges to accurate detection. Under the MIL paradigm, these scale disparities intensify the inconsistency between the confidence scores of the predicted bounding boxes and their actual positional accuracy, leading to suboptimal detection performance. To mitigate this issue, we propose the scale augmentation module, which is structured around the design of SSC loss.

In an ideal scenario, the predicted size of the same object, represented by the scale of the proposal with the highest confidence score, should remain consistent across views with different resolutions. To enforce this consistency, the SSC loss is designed to minimize the distributional disparity in predicted confidence scores between the original and resized views.

\begin{figure}[!tb]
    \centering
    \includegraphics[width=1\textwidth]{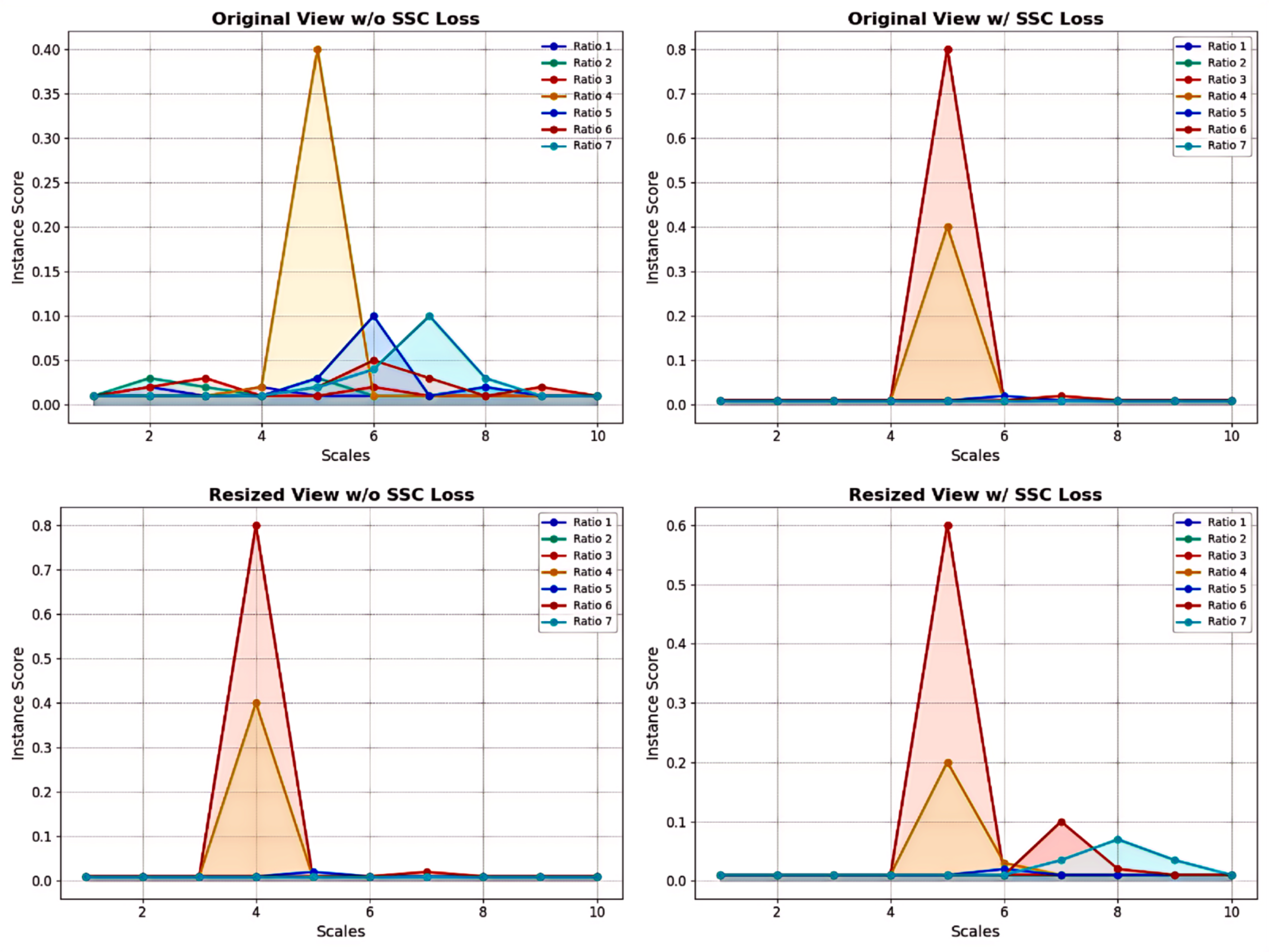}
    \caption{Line graphs of instance score distributions from the original and the resized views before and after employing the proposed SSC loss, the distributions are grouped by ratios.}
    \label{fig:SSC}
\end{figure}

Starting with the proposal bags \( B_o \) derived from the original view, we obtain the corresponding resized proposal bags \( B_d \) from the resized view. The output scores, including class scores and instance scores of \(B_o\) and \(B_d\), are generated through the dual-stream branches integrated within the classical MIL head. For the \(i\)-th proposal bag \( B_i \), the scores obtained from the original view are denoted as $S_{\text{i$_o$}}^{cls}$ and $S_{\text{i$_o$}}^{ins}$ respectively, while the corresponding scores from the resized view are represented as $S_{\text{i$_d$}}^{cls}$ and $S_{\text{i$_o$}}^{ins}$. These sets of scores have been processed through an activation function (e.g., softmax), and their dimensions are \( \mathbb{R}^{N \times C} \), where 
\( N \) denotes the total number of proposals within the bag and  \( C \) represents the number of object categories being classified. To maintain consistency in scale across the outputs from different views, the proposals' score distributions are initially adjusted according to their respective scales. In detail, a set of fundamental scales, denoted as \(\{s_1, s_2, \ldots, s_G\}\), is established, where \(G\) represents the total number of these predefined scales. The output score dimensions are then transformed from \(\mathbb{R}^{N \times C}\) to \(\mathbb{R}^{G \times K}\), where \(K\) accounts for scale-independent variables like category and aspect ratio. Once the score distributions are adjusted according to the scale for each view, cosine similarity is utilized to evaluate the consistency between them:
\begin{equation}
\textit{sim}_{m,g}^{\textit{ins}} = 1 - \frac{\left[S_{\textit{i$_{o}$}}^{\textit{ins}}\right]_{m,g} \cdot \left[S_{\textit{i$_{d}$}}^{\textit{ins}}\right]_{m,g}}{\left\|\left[S_{\text{i$_{o}$}}^{\textit{ins}}\right]_{m,g}\right\| \cdot \left\|\left[S_{\textit{i$_{d}$}}^{\text{ins}}\right]_{m,g}\right\|},
\label{eq:sim1}
\end{equation}
\begin{equation}
\textit{sim}_{m,g}^{\textit{cls}} = 1 - \frac{\left[S_{\textit{i$_{o}$}}^{\textit{cls}}\right]_{m,g} \cdot \left[S_{\textit{i$_{d}$}}^{\textit{cls}}\right]_{m,g}}{\left\|\left[S_{\textit{i$_{o}$}}^{\textit{cls}}\right]_{m,g}\right\| \cdot \left\|\left[S_{\textit{i$_{d}$}}^{\textit{cls}}\right]_{m,g}\right\|},
\label{eq:sim2}
\end{equation}
where \(m\) refers to the \(m\)-th point label, while \(g\) denotes the \(g\)-th group of proposals, categorized according to the predefined basic scales. Utilizing these similarity measurements, the overall SSC loss is expressed as follows:
\begin{equation}
\mathcal{L}_{\textit{SSC}} = \sum_{m=1}^{M} \sum_{g=1}^{G} \left\{ \omega_1 \ell_s \left( \textit{sim}_{m,g}^{\textit{ins}}, 0 \right) + \omega_2 \ell_s \left( \textit{sim}_{m,g}^{\textit{cls}}, 0 \right) \right\},
\label{eq:SSC}
\end{equation}
where \(M\) represents the total number of point labels, \(\ell_s\) denotes the SmoothL1 loss function, and \(\omega_1\) and \(\omega_2\) are the weights assigned to the loss terms, with values of 2.0 and 1.0, respectively. By incorporating the SSC loss, the MIL network ensures alignment of the score distributions for proposals associated with the same label across multiple views, as illustrated in Fig.~\ref{fig:SSC}. This alignment effectively reduces discrepancies between confidence scores and positional accuracy, thereby improving the precision in perceiving object scale.

\subsubsection{Scale-Sensitive Feature Fusion Module}\label{subsubsec4}

As mentioned in Sec. \ref{sec1}, in traditional MIL-based methods for extracting features from ROIs, a relatively rigid feature layer assignment strategy is often employed, heavily relying on the scale information of the proposals and predefined hyperparameters. To be specific, assuming the feature pyramid is \(\textit{lvls}_{\text{fea}} = \{p_1, p_2, \ldots, p_P\}\) with \(P\) levels, ROIs are also assigned to different levels based on their scale, calculated as \(\textit{lvls}_{\textit{ROI}} = \log_2\left(\sqrt{w \cdot h} / \textit{fs} + 1 \times 10^{-6}\right)\), where \(\textit{fs}\) is a predefined scale parameter finest scale, and \(w\) and \(h\) represent the ROI's width and height, respectively. ROI features are then extracted from the corresponding feature layers. However, this approach to feature layer mapping can lack flexibility, particularly when the ROI's scale is not precise, which may result in incorrect layer assignments. Additionally, for ROIs with dimensions close to the boundaries of different layers, the feature layer mapping might become unstable, negatively affecting the consistency of feature extraction.

To address this problem, we devise the SSFF module for the MIL head, as shown in the below part of Fig.~\ref{fig:main_method}. Drawing inspiration from the MoE architecture, each feature layer undergoes a gating mechanism during training to produce its corresponding gating score. These scores are subsequently utilized to automatically combine the multiple output layers of the FPN. The final fused feature map is then used for feature extraction. Specifically, the multiple output layers of the FPN are automatically aggregated based on a self-activated gating score: 
\begin{equation}
G_n = \text{softmax}\left(\text{conv}\left(F_n\right)\right),
\label{eq:gate_score}
\end{equation}
where \(F_n\) represents the \(n\)-th FPN feature layer, which is scaled to a unified shape using nearest-neighbor interpolation. \(G_n\) denotes the gating score for each layer; \(\text{conv}(\cdot)\) is a \(3 \times 3\) convolutional layer with a single output channel; and \(\text{softmax}(\cdot)\) normalizes the sum of \(G_1, G_2, \ldots, G_N\) to one for each pixel.

The final one-layer feature map F can be obtained by:
\begin{equation}
F = \sum_{n=1}^{N} G_n F_n,
\label{eq:single_layer_feature}
\end{equation}
where \(N\) denotes the total number of layers used in the FPN, and \(N = 4\) by default, corresponding to the $P_2$ to $P_5$ layers. In order to sum up all the feature maps together, we use interpolation to resize the $P_3$ to $P_5$ feature maps to match the size of the $P_2$ feature map. Eqs. (\ref{eq:gate_score}) and (\ref{eq:single_layer_feature}) demonstrate that the weights used to fuse the layers are dynamically generated by the layers themselves.

Such a design can automatically deal with ROIs of different sizes. Most importantly, this module ensures that the final extracted ROI features contain more accurate scale information. Adopting this more flexible method helps mitigate issues such as mapping deviation and inconsistencies in feature extraction caused by inaccuracies in proposal scales. Taking advantage of this novel mechanism SSFF, PointOBB-v3 achieves a much higher AP$_{50}$ compared to our conference work PointOBB~\cite{luo2024pointobb} (40.18\% vs 37.31\% on DIOR-R testing set and 50.44\% vs 33.31\% on DOTA-v1.0 testing set, see in Tab. \ref{tab:SSC&DS&SSFF}).

\subsection{Angle Acquisition Module}\label{subsec7}
To enable orientation learning without direct angle supervision, we start by taking into account the natural symmetry properties of objects, which, as discussed earlier in Sec. \ref{subsec4}, play a crucial role in understanding object orientation under weakly supervised scenarios. Previous research has examined symmetry within the context of HBox supervision \cite{Yu_Yang_Li_Zhou_Zhang_Yan_Da_2023}. Our findings reveal that symmetry-based self-supervised learning demonstrates resilience to annotation noise. This suggests that it is feasible to achieve accurate angle prediction even with only a single point annotation.

\subsubsection{Dense-to-Dense Assignment}\label{subsubsec5}
As outlined in Sec. \ref{subsubsec1}, during the second stage, an angle acquisition module is developed based on the rotated/flipped view, incorporating a self-supervised angle branch to facilitate angle learning. Both views are passed through feature extractors with shared parameters, such as ResNet50 \cite{He_Zhang_Ren_Sun_2016} and FPN \cite{Lin_Dollar_Girshick_He_Hariharan_Belongie_2017}, to generate dense feature pyramid representations. In the absence of scale information, grid points across all feature levels within the central region surrounding the ground-truth points are selected as positive samples. For positive samples associated with the same point label at a given level, their predicted angles are averaged to derive the final prediction value.

\subsubsection{Dense-to-Sparse Matching}\label{subsubsec6}
When matching dense feature-based angle predictions with sparse feature-based proposals, directly searching for the nearest grid points to a proposal's center is not an appropriate approach. Because of the possible disparity between the receptive field of the grid points and the scale of the object proposals, the angle predictions might not accurately correspond to the actual object region. To address this issue, we implement a hierarchical pairing strategy, ensuring a consistent alignment between the receptive fields used for angle prediction and the scale of the proposals. Given a feature pyramid denoted as \(\textit{lvls}_{\text{fea}} = \{p_1, p_2, \ldots, p_P\}\) with \(P\) levels, proposals are categorized into the level \(\textit{lvl}_{\text{prop}}\) based on their scale, aligning with the ROI assignment strategy used in two-stage object detection algorithms like Oriented R-CNN~\cite{Xie_Cheng_Wang_Yao_Han_2021}. The orientation of each proposal is determined using the average angle predictions from the central region of the proposal on the feature map \(\textit{lvl}_{\text{prop}}\), as illustrated in Fig.~\ref{fig:main_method}. This DS matching strategy can effectively aggregate the dense angle predictions to correspond to the sparse object proposals.

\subsubsection{Self-Supervised Angle Loss}\label{subsubsec7}
Leveraging the affine transformation relationship between the original view and the rotated/flipped view, object angles are learned using the Self-Supervised Angle (SSA) loss.
If the enhanced view is created through a random rotation by an angle \(\theta'\), the angle predictions from both the original and rotated views should align with and uphold the same rotational relationship. When the enhanced view is produced through a vertical flip, the angle predictions must account for differences of \(k\pi\), where \(k\) is an integer ensuring the results remain within the same periodic cycle. The loss between the outputs of the two views can be formulated as:
\begin{equation}
\begin{cases}
\mathcal{L}_{\text{rot}} = \min_{k \in \mathbb{Z}} \sum_{p=1}^{P} \ell_{\text{angle}} \left( \theta_{\text{rot}}^p - \theta^p, k\pi + \theta' \right) \\
\mathcal{L}_{\text{flip}} = \min_{k \in \mathbb{Z}} \sum_{p=1}^{P} \ell_{\text{angle}} \left( \theta_{\text{flip}}^p + \theta^p, k\pi \right)
\end{cases},
\label{eq:SSA_1}
\end{equation}
where \(\ell_{\text{angle}}\) refers to the SmoothL1 loss, while \(\theta^p\), \(\theta_{\text{flip}}^p\), and \(\theta_{\text{rot}}^p\) denote the angle predictions at the \(p\)-th level of the feature pyramids for the original view, the rotated/flipped view by flipping, and the rotated/flipped view by rotation, respectively. And the SSA loss is represented as:
\begin{equation}
\mathcal{L}_{\text{SSA}} = \mathcal{L}_{\text{rot}} + \mathcal{L}_{\text{flip}}.
\label{eq:SSA_2}
\end{equation}

\subsection{End-to-End Framework Design for PointOBB}\label{subsec8}
\subsubsection{Pipeline}\label{subsubsec8}
\begin{figure}[!tb]
    \centering
    \includegraphics[width=1\textwidth]{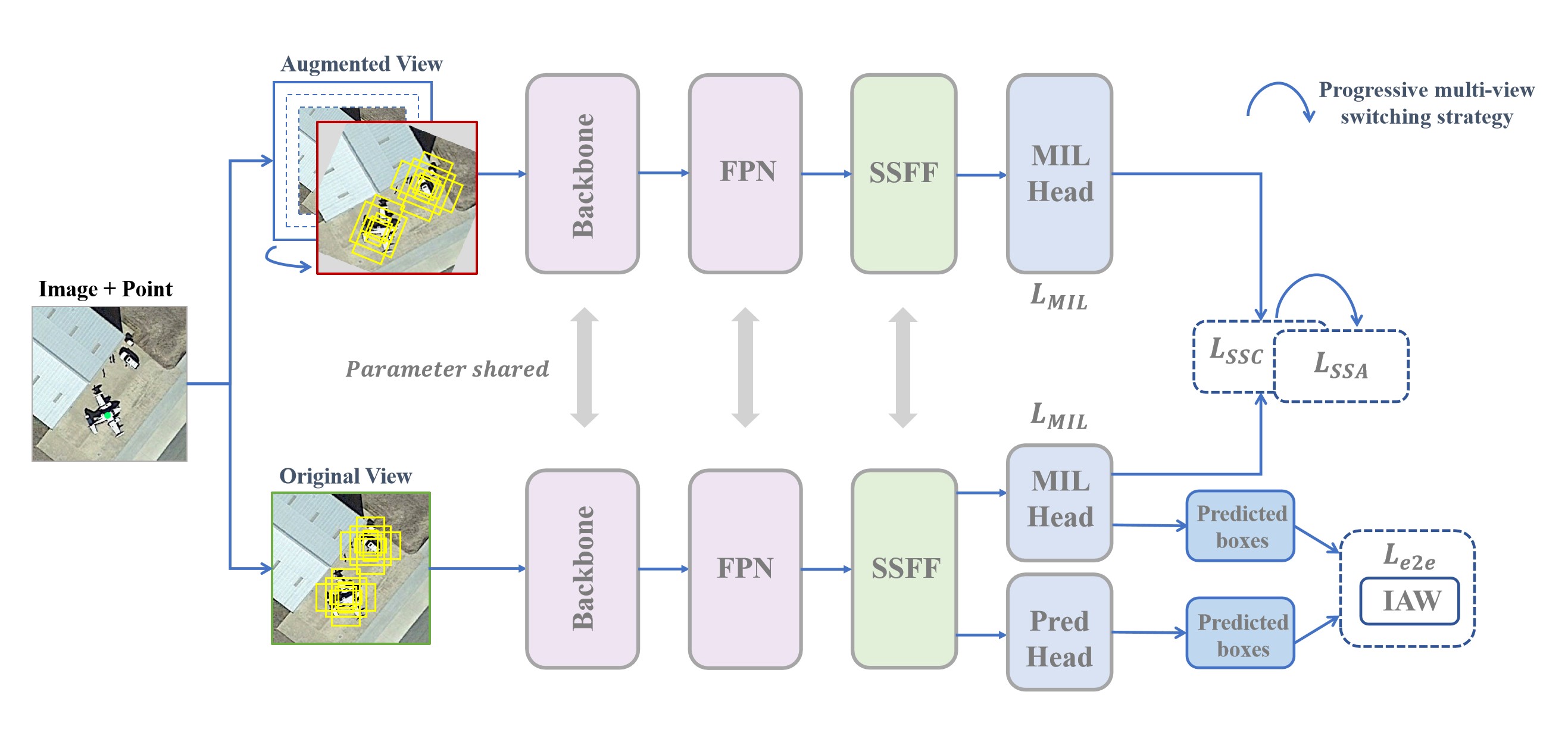}
    \caption{The Pipeline of End-to-end version of PointOBB-v3. Building on the original MIL-based framework, a detection branch is integrated. This newly added branch shares the backbone and neck parameters with the MIL head, promoting parameter efficiency. To facilitate the joint training of both branches, the outputs of the MIL branch and the detection branch are aligned through the calculation of L$_{e2e}$. Furthermore, an Instance-Aware Weighting (IAW) strategy is proposed to further enhance the integration and collaborative training of the two branches.}
    \label{fig:e2e}
\end{figure}

Our end-to-end framework is illustrated in Fig.~\ref{fig:e2e}. Building upon the original MIL-based framework, we introduced a detection branch utilizing a Rotated FCOS head \cite{Tian_Shen_Chen_He_2019}. This new branch shares the backbone and neck parameters with the MIL branch, ensuring parameter efficiency. To enable joint training of the two branches, we align the outputs of the MIL head and the Rotated FCOS head by calculating $\mathcal{L}_{class}$, $\mathcal{L}_{box}$, and $\mathcal{L}_{ctr}$, which represent the classification loss, bounding box regression loss, and the centerness loss, respectively.

Moreover, to address the adverse impact of low-quality predictions from the MIL head on the detection branch during joint training, we propose an IAW strategy. This strategy dynamically assigns smaller weights to lower-quality predictions from the MIL head during loss computation, thereby reducing their influence on parameter optimization. This mechanism effectively filters out poor-quality predictions, ensuring better alignment and improving the overall training process, and we will elaborate on it further in the next section.

With the above pipeline design, the model can be trained and inferred in a more convenient and time-efficient end-to-end manner. As demonstrated in Tab. \ref{tab:epoch} , compared to the two-stage version, the end-to-end version reduces the training hours by approximately 21.36\%, while maintaining competitive detection performance.

\subsubsection{Instance-Aware Weighting Strategy}\label{subsubsec9}
During the joint training process of the two branches, the predictions from the MIL head vary significantly in quality, especially in the early stages of training, where low-quality predictions dominate. Aligning the two branches using these low-quality predictions would inevitably have a detrimental effect on parameter optimization. Clearly, assigning the same weight to all instances during loss computation is not an appropriate approach. Intuitively, we aim to assign smaller weights to lower-quality predictions during loss calculation to reduce their negative impact on optimization.

Based on this analysis, we propose the IAW strategy. Specifically, for each instance, we calculate a dynamic weight $w_j$ by multiplying its class score with its instance score, and the final loss is obtained by summing the weighted losses across all instances, which can be formed as:
\begin{equation}
\mathcal{L}_{e2e} = \mathcal{L}_{{class}} + \sum_{j} \text{w}_j \cdot \left(\mathcal{L}_{{box}_j} + \mathcal{L}_{{ctr}_j} \right),
\label{eq: loss_e2e}
\end{equation}
where $w_j$ = $S_j^{cls} \odot S_j^{ins}$, $j$ means the $j_{th}$ instance. Under this dynamic weighting strategy, lower-quality predictions, which tend to have lower classification and instance scores, will contribute less to the final loss. This achieves the goal of filtering and screening out poor-quality predictions, enhancing the overall joint training process.

\subsection{Optimization with Overall MIL Loss}\label{subsec9}
Apart from the previously introduced SSC and SSA losses, we also incorporate the MIL loss within our approach. As illustrated in Fig.~\ref{fig:main_method}, each constructed view is processed through the MIL head, while the output from the original view passes through an additional refined MIL head to produce the final predictions. The associated loss can be summarized as follows.

In detail, each point label corresponds to a specific proposal bag. Using the MIL head, the instance score and class score for the \(n\)-th proposal within the \(i\)-th proposal bag are represented as \(S_{\textit{i,n}}^{ins}\) and \(S_{\textit{i,n}}^{cls}\), respectively. The overall output for the entire proposal bag is expressed as \(S_i = \sum_{n=1}^{N} S_{\textit{i,n}}^{ins} \odot S_{\textit{i,n}}^{cls}\), where \(N\) indicates the total number of proposals in the bag. The MIL losses corresponding to the original view, rotated/flipped view, and resized view are represented as $\mathcal{L}_{\text{MIL}}^{\text{ori}}$, $\mathcal{L}_{\text{MIL}}^{\text{rfv}}$, and $\mathcal{L}_{\text{MIL}}^{\text{res}}$, respectively. These three losses follow the common form of the general MIL loss $\mathcal{L}_{\text{MIL}}^{\text{init}}$, which is defined as:
\begin{equation}
\mathcal{L}_{\text{MIL}}^{\text{init}} = -\frac{1}{I} \sum_{i=1}^{I} \sum_{c=1}^{C} \left\{ Q_{i,c} \log(S_{i,c}) + (1 - Q_{i,c})(1 - \log(1 - S_{i,c})) \right\},
\label{eq:MIL_init}
\end{equation}
where \(I\) denotes the number of proposal bags in the batch, \(C\) represents the total number of categories, \(Q_{i,c}\) is the one-hot encoded category label, and \(S_{i,c}\) corresponds to the score for the \(c\)-th category in \(S_i\). For the refined MIL head, $\mathcal{L}_{\text{MIL}}^{\text{ref}}$ utilizes focal loss \cite{Lin_Goyal_Girshick_He_Dollar_2017} to compute the classification loss between \(Q_{i,c}\) and \(S_{i,c}\). Therefore, considering the proposed progressive multi-view switching strategy is formulated as:
\begin{equation}
\mathcal{L}_{\text{MIL}} = \mathcal{L}_{\text{MIL}}^{\text{ori}} + \mathcal{L}_{\text{MIL}}^{\text{ref}} + \alpha \mathcal{L}_{\text{MIL}}^{\text{rfv}} + \beta \mathcal{L}_{\text{MIL}}^{\text{res}},
\label{eq:MIL}
\end{equation}
where \(\alpha\) is set to 0 and \(\beta\) to 1 during the first stage, while \(\alpha\) is set to 1 and \(\beta\) to 0 in the second and third stages, as described in Sec. \ref{subsubsec2}. The total loss of our framework is formulated as:
\begin{equation}
\mathcal{L} = \mathcal{L}_{\text{MIL}} + \alpha \mathcal{L}_{\text{SSC}} + \beta \mathcal{L}_{\text{SSA}},
\label{eq:LOSS_all}
\end{equation}
where $\alpha$ and $\beta$ keep the same setting with Eq. \ref{eq:LOSS_all}.

For the end-to-end version, the total loss is similarly formulated as:
\begin{equation}
\mathcal{L} = \mathcal{L}_{\text{MIL}} + \alpha \mathcal{L}_{\text{SSC}} + \beta \mathcal{L}_{\text{SSA}} + \gamma \mathcal{L}_{e2e},
\label{eq:LOSS_all_e2e}
\end{equation}
where $\gamma$ is a hyperparameter and is set to \(1\) and $\alpha$/$\beta$ also keeps the same setting with Eq. \ref{eq:LOSS_all}.

\section{Experiments}\label{sec5}
\subsection{Datasets and Implementation Details}\label{subsec10}
\noindent\textbf{Dataset.} DIOR-R \cite{Cheng_2022} is an updated aerial object detection dataset with OBB annotations, derived from its predecessor DIOR \cite{Li_Wan_Cheng_Meng_Han_2020}, and includes 23,463 images, spanning 20 categories and 190,288 sizes. DOTA-v1.0 \cite{Xia_Bai_Ding_Zhu_Belongie_Luo_Datcu_Pelillo_Zhang_2018} consists of 2,806 images containing 188,282 instances across 15 categories. The image resolutions range from 800×800 to 4,000×4,000 pixels, demonstrating notable variations in scale and orientation. DOTA-v1.5 builds upon DOTA-v1.0 by incorporating annotations for very small objects (smaller than 10 pixels) and adding a new category container crane. This version includes 403,318 instances while maintaining the same number of images and dataset split as DOTA-v1.0. DOTA-v2.0 further extends the dataset to 11,268 images and 1,793,658 instances across 18 categories, introducing two new categories, airport and helipad, to enhance diversity and difficulty. FAIR1M \cite{sun2021fair1mbenchmarkdatasetfinegrained} is a remote sensing dataset containing over 1 million instances and more than 40,000 images, designed for fine-grained object recognition. The dataset includes annotations across 37 categories, with results evaluated on the FAIR1M-1.0 server. STAR \cite{STAR} is a comprehensive dataset for scene graph generation, encompassing over 210,000 objects with varying spatial resolutions. It includes 48 fine-grained categories and features precise annotations using oriented bounding boxes. RSAR \cite{zhang2025rsar} is a comprehensive multi-class large-scale rotated SAR object detection dataset, which comprises 95,842 SAR images and 183,534 annotated instances across 6 typical SAR object categories.

\noindent\textbf{Single point annotation.}
To effectively replicate the biases present in manual annotation, point labels are generated based on the central region of the OBB annotations. PointOBB \cite{luo2024pointobb} has demonstrated through extensive ablation experiments that introducing appropriate noise is more effective than using center point labels. This paper directly adopts its default optimal configuration. A random point is selected within a range defined by 10\% of the OBB's height and width to serve as the label.

\noindent\textbf{Experiment settings.}
The algorithms used in our experiments are sourced from two open-source PyTorch-based libraries: MMRotate \cite{Zhou_2022} and MMDetection \cite{chen2019mmdetectionopenmmlabdetection}. In accordance with the default settings in MMRotate, large images in the DOTA-v1.0 dataset are cropped into 1,024 × 1,024 patches with a 200-pixel overlap. For the DIOR-R dataset, the original image size of 800 × 800 is retained. 

Experiments are performed on a server with 2 A100 GPUs and 80GB memory. For the two-stage version, we use the ``2×" schedule to train all methods and apply the ``1×" schedule when training RBox-supervised algorithms with our generated pseudo OBBs. For the end-to-end version, the entire detector is trained using the ``2×" schedule. This paper inherits PointOBB's optimal default settings, employing the SGD optimizer with a learning rate of 0.005, momentum of 0.9, and weight decay of 0.0001. A linear warm-up strategy is applied for the initial 500 iterations with a rate of 0.001 and batch size is 2. With the exception of WSODet \cite{Tan_Jiang_Guo_Zhang_2023}, which utilizes VGG16 \cite{Girshick_Donahue_Darrell_Malik_2014}, all the listed models are built on the ResNet50 \cite{He_Zhang_Ren_Sun_2016} backbone, pre-trained on ImageNet \cite{imagenet}. For the comparative algorithms, their default settings are adopted. In our method, for both the two-stage version and the end-to-end version, the finest scale is set to 56, and the scale factor \(\sigma\) for the resized view is randomly selected between 0.5 and 1.5. Following the results of the ablation study, ``burn-in step1" and ``burn-in step2" are configured at the 6th and 8th epochs, respectively. Additionally, we adopt scale-based grouping, which has been demonstrated through experiments in PointOBB to be more effective than proposal-based and ratio-based grouping strategies.

\noindent\textbf{Evaluation metric.}
Mean Average Precision (mAP) is used as the main metric to evaluate and compare our methods with existing approaches. Additionally, to more effectively evaluate the quality of pseudo labels derived from points, we report the mean Intersection over Union (mIoU) between the ground-truth boxes and the predicted pseudo OBBs on the training set.

\begin{figure}[h]
    \centering
    \includegraphics[width=1\textwidth]{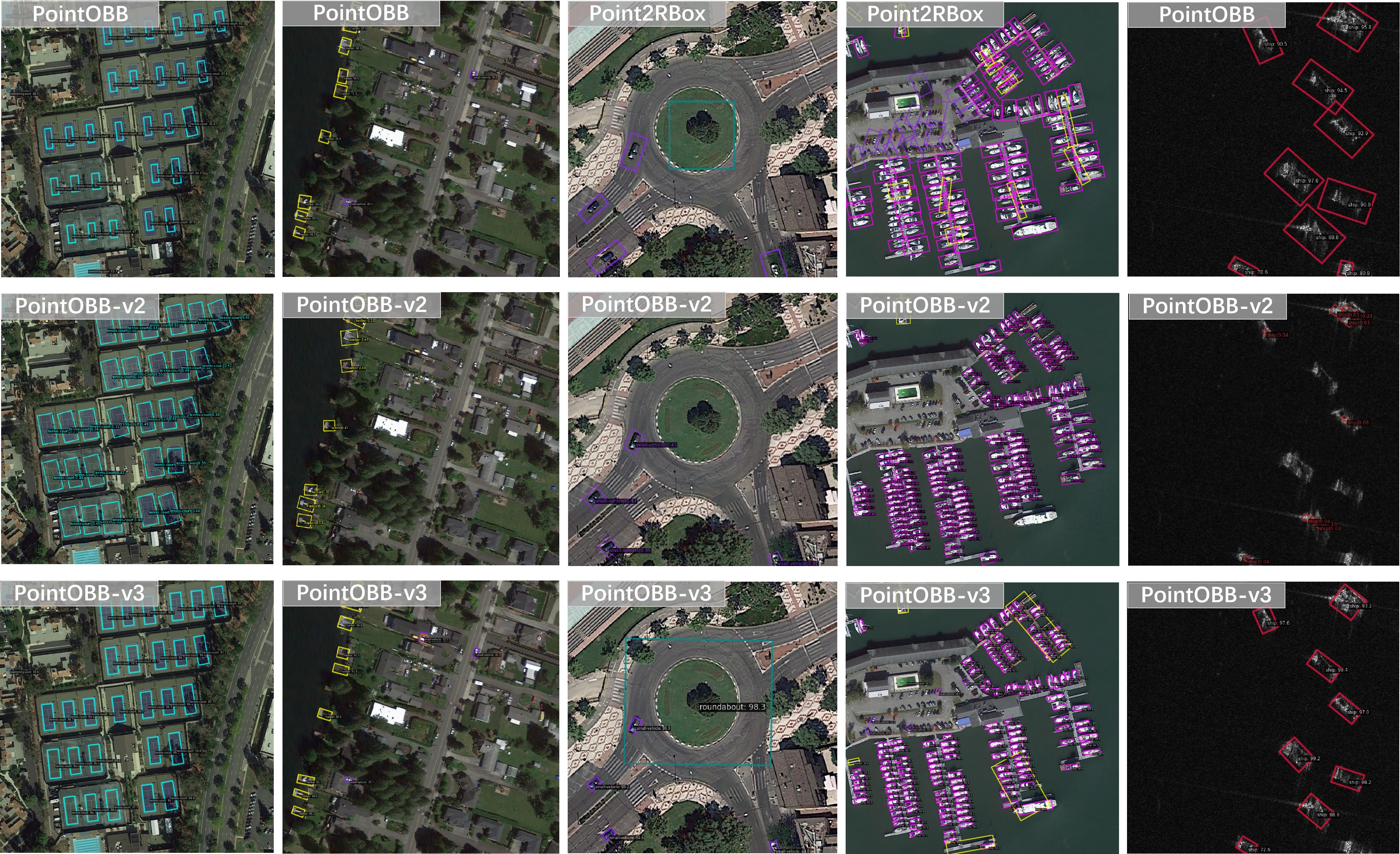}
    \caption{The visual detection results. Compared existing methods include: PointOBB (2024) \cite{luo2024pointobb}, PointOBB-v2 (2024) \cite{ren2024pointobbv2simplerfasterstronger}, and Point2RBox (2024) \cite{yu2024point2rbox}. The last row showcases the results of the proposed PointOBB-v3 combined with Oriented R-CNN.}
    \label{fig:show}
\end{figure}

\begin{sidewaystable}
\centering
\begin{minipage}[t]{0.48\textwidth}
\captionsetup{justification=raggedright, singlelinecheck=false}
\caption{Accuracy on the DIOR-R testing set.}\label{tab:main_dior}
\tiny
\setlength{\tabcolsep}{2pt}
\renewcommand{\arraystretch}{0.8}
\begin{tabular*}{\textheight}{@{\extracolsep\fill}lcccccccccccccccccccccccc}
\toprule
\textbf{Method} & \textbf{*} & \textbf{APL} & \textbf{APO} & \textbf{BF} & \textbf{BC} & \textbf{BR} & \textbf{CH} & \textbf{ESA} & \textbf{ETS} & \textbf{DAM} & \textbf{GF} & \textbf{GTF} & \textbf{HA} & \textbf{OP} & \textbf{SH} & \textbf{STA} & \textbf{STO} & \textbf{TC} & \textbf{TS} & \textbf{VE} & \textbf{WM} & \textbf{mAP\textsubscript{50}}\\
\midrule
\multicolumn{23}{@{}l}{\textbf{RBox-supervised:}} \\
Rotated RetinaNet \cite{Lin_Goyal_Girshick_He_Dollar_2017} & \checkmark & 58.9 & 19.8 & 73.1 & 81.3 & 17.0 & 72.6 & 68.0 & 47.3 & 20.7 & 74.0 & 73.9 & 32.5 & 32.4 & 75.1 & 67.2 & 58.9 & 81.0 & 44.5 & 38.3 & 62.6 & 54.96 \\
Rotated FCOS \cite{Tian_Shen_Chen_He_2019} & \checkmark & 61.4 & 38.7 & 74.3 & 81.1 & 30.9 & 72.0 & 74.1 & 62.0 & 25.3 & 69.7 & 79.0 & 32.8 & 48.5 & 80.0 & 63.9 & 68.2 & 81.4 & 46.4 & 42.7 & 64.4 & 59.83 \\
Oriented R-CNN \cite{Xie_Cheng_Wang_Yao_Han_2021} & \checkmark & 63.1 & 34.0 & 79.1 & 87.6 & 41.2 & 72.6 & 76.6 & 65.0 & 26.9 & 69.4 & 82.8 & 40.7 & 55.9 & 81.1 & 72.9 & 62.7 & 81.4 & 53.6 & 43.2 & 65.6 & \textbf{62.80} \\
\midrule
\multicolumn{23}{@{}l}{\textbf{Image-supervised:}} \\
WSODet \cite{Tan_Jiang_Guo_Zhang_2023} & \checkmark & 20.7 & 29.0 & 63.2 & 67.3 & 0.2 & 65.5 & 0.4 & 0.1 & 0.3 & 49.0 & 28.9 & 0.3 & 1.5 & 1.2 & 53.4 & 16.4 & 40.0 & 0.1 & 6.1 & 0.1 & 22.20 & \\
\midrule
\multicolumn{23}{@{}l}{\textbf{HBox-supervised:}} \\
H2RBox \cite{Yang_Zhang_Li_Wang_Zhou_Yan_2022} & \checkmark & 68.1 & 13.0 & 75.0 & 85.4 & 19.4 & 72.1 & 64.4 & 60.0 & 23.6 & 68.9 & 78.4 & 34.7 & 44.2 & 79.3 & 65.2 & 69.1 & 81.5 & 53.0 & 40.0 & 61.5 &  \textbf{57.80} \\
H2RBox-v2 \cite{Yu_Yang_Li_Zhou_Zhang_Yan_Da_2023} & \checkmark & 67.2 & 37.7 & 55.6 & 80.8 & 29.3 & 66.8 & 76.1 & 58.4 & 26.4 & 53.9 & 80.3 & 25.3 & 48.9 & 78.8 & 67.6 & 62.4 & 82.5 & 49.7 & 42.0 & 63.1 & 57.64 \\
\midrule
\multicolumn{23}{@{}l}{\textbf{Point-supervised:}} \\
Point2Mask-RBox \cite{Li_Yuan_Wang_Zhu_Li_Liu_Zhang} & $\times$ & 15.6 & 0.1 & 50.6 & 25.4 & 4.2 & 50.9 & 23.8 & 17.5 & 8.1 & 0.8 & 9.6 & 1.4 & 15.3 & 1.6 & 5.8 & 6.3 & 17.9 & 7.1 & 4.5 & 9.2 & 13.77 \\
P2BNet \cite{Chen_Yu_Han_Hassan_Wang_Li_Zhao_Shi_Han_Ye_2022} + H2RBox \cite{Yang_Zhang_Li_Wang_Zhou_Yan_2022} & $\times$ & 52.7 & 0.1 & 60.6 & 80.0 & 0.1 & 22.6 & 11.5 & 5.2 & 0.7 & 0.2 & 42.8 & 2.8 & 0.2 & 25.1 & 8.6 & 29.1 & 69.8 & 9.6 & 7.4 & 22.6 & 22.59 \\
P2BNet \cite{Chen_Yu_Han_Hassan_Wang_Li_Zhao_Shi_Han_Ye_2022} + H2RBox-v2 \cite{Yu_Yang_Li_Zhou_Zhang_Yan_Da_2023} & $\times$ & 51.6 & 3.0 & 65.2 & 78.3 & 0.1 & 8.1 & 7.6 & 6.3 & 0.8 & 0.3 & 44.9 & 2.3 & 0.1 & 35.9 & 9.3 & 39.2 & 79.0 & 8.8 & 10.3 & 21.3 & 23.61 \\
Point2RBox \cite{yu2024point2rbox} & \checkmark & 42.0 & 9.1 & 62.9 & 52.8 & 10.8 & 72.2 & 3.0 & 43.9 & 5.5 & 9.7 & 25.1 & 9.1 & 21.0 & 24.0 & 20.4 & 25.1 & 71.7 & 4.5 & 16.1 & 16.3 & 27.34 \\
PointOBB (FCOSR) \cite{luo2024pointobb} & $\times$ & 58.4 & 17.1 & 70.7 & 77.7 & 0.1 & 70.3 & 64.7 & 4.5 & 7.2 & 0.8 & 74.2 & 9.9 & 9.1 & 69.0 & 38.2 & 49.8 & 46.1 & 16.8 & 32.4 & 29.6 & 37.31 \\
PointOBB (ORCNN) \cite{luo2024pointobb} & $\times$ & 58.2 & 15.3 & 70.5 & 78.6 & 0.1 & 72.2 & 69.6 & 1.8 & 3.7 & 0.3 & 77.3 & 16.7 & 4.0 & 79.2 & 39.6 & 51.7 & 44.9 & 16.8 & 33.6 & 27.7 & 38.08 \\
PointOBB-v2 (FCOSR) \cite{ren2024pointobbv2simplerfasterstronger} & $\times$ & 45.8 & 6.4 & 58.2 & 66.5 & 2.4 & 60.9 & 40.3 & 33.2 & 4.5 & 55.3 & 34.2 & 11.4 & 8.9 & 43.0 & 46.8 & 37.3 & 79.6 & 21.9 & 10.6 & 19.8 & 34.35 \\
PointOBB-v2 (ORCNN) \cite{ren2024pointobbv2simplerfasterstronger}  & $\times$ & 41.5 & 6.0 & 59.7 & 79.7 & 6.7 & 72.2 & 50.0 & 41.0 & 5.8 & 59.4 & 38.3 & 11.2 & 21.6 & 53.0 & 54.4 & 49.3 & 80.9 & 33.2 & 11.1 & 17.2 & 39.62 \\
\rowcolor{gray!30}
PointOBB-v3 (e2e) & \checkmark & 53.1 & 11.0 & 70.4 & 79.0 & 0.1 & 70.1 & 53.0 & 1.6 & 0.9 & 45.7 & 69.4 & 3.1 & 3.4 & 63.2 & 44.5 & 57.3 & 58.1 & 1.3 & 30.3 & 36.5 & 37.60 \\
\rowcolor{gray!30}
PointOBB-v3 (FCOSR) & $\times$ & 37.1 & 13.0 & 70.2 & 79.4 & 0.1 & 71.5 & 44.7 & 50.8 & 4.5 & 52.9 & 74.7 & 2.5 & 0.5 & 69.6 & 46.1 & 58.0 & 58.9 & 3.0 & 31.8 & 34.4 & 40.18 \\
\rowcolor{gray!30} 
PointOBB-v3 (ORCNN) & $\times$ & 32.8 & 11.2 & 69.9 & 80.3 & 0.1 & 72.3 & 42.1 & 55.3 & 0.7 & 54.8 & 77.9 & 1.9 & 0.8 & 79.8 & 53.3 & 60.5 & 68.1 & 0.9 & 33.2 & 40.5 & \textbf{41.82} \\
\botrule
\end{tabular*}
\footnotetext{The categories in DIOR-R include Airplane (APL), Airport (APO), Baseball Field (BF), Basketball Court (BC), Bridge (BR), Chimney (CH), Expressway Service Area (ESA), Expressway Toll Station (ETS), Dam (DAM), Golf Field (GF), Ground Track Field (GTF), Harbor (HA), Overpass (OP), Ship (SH), Stadium (STA), Storage Tank (STO), Tennis Court (TC), Train Station (TS), Vehicle (VE) and Windmill (WM). "-RBox" means the minimum rectangle operation is performed on the Mask to obtain RBox. \textbf{*} Comparison tracks: \checkmark = End-to-end training and testing; $\times$ = Generating pseudo labels to train the detector (two-stage training).}
\end{minipage}%
\hfill
\begin{minipage}[t]{0.48\textwidth}
\captionsetup{justification=raggedright, singlelinecheck=false}
\caption{Accuracy on the DOTA-v1.0 testing set.}\label{tab:main_dota10}
\tiny
\setlength{\tabcolsep}{2pt}
\renewcommand{\arraystretch}{0.8}
\begin{tabular*}{\textheight}{@{\extracolsep\fill}lccccccccccccccccc}
\toprule
\textbf{Method} & \textbf{*} & \textbf{PL} & \textbf{BD} & \textbf{BR} & \textbf{GTF} & \textbf{SV} & \textbf{LV} & \textbf{SH} & \textbf{TC} & \textbf{BC} & \textbf{ST} & \textbf{SBF} & \textbf{RA} & \textbf{HA} & \textbf{SP} & \textbf{HC} & \textbf{mAP\textsubscript{50}} \\
\midrule
\multicolumn{17}{@{}l}{\textbf{RBox-supervised:}} \\
Rotated RetinaNet \cite{Lin_Goyal_Girshick_He_Dollar_2017} & \checkmark & 88.7 & 77.6 & 38.8 & 58.2 & 74.6 & 71.6 & 79.1 & 88.0 & 80.2 & 72.3 & 52.8 & 58.6 & 62.6 & 67.7 & 59.6 & 68.69 \\
Rotated FCOS \cite{Tian_Shen_Chen_He_2019} & \checkmark & 88.4 & 76.8 & 45.0 & 59.2 & 79.2 & 79.0 & 86.9 & 88.1 & 76.6 & 78.8 & 58.6 & 57.5 & 69.3 & 72.4 & 53.5 & \textbf{71.28} \\
\midrule
\multicolumn{17}{@{}l}{\textbf{HBox-supervised:}} \\
Sun et al. \cite{Sun_Ran_Yang_Gao_Kurozumi_Kimata_Ye_2021} & $\times$ & 51.5 & 38.7 & 16.1 & 36.8 & 29.8 & 19.2 & 23.4 & 83.9 & 50.6 & 80.0 & 18.9 & 50.2 & 25.6 & 28.7 & 25.5 & 38.60 \\
H2RBox \cite{Yang_Zhang_Li_Wang_Zhou_Yan_2022} & \checkmark & 88.5 & 73.5 & 48.8 & 56.9 & 77.5 & 65.4 & 77.9 & 88.9 & 81.2 & 79.2 & 55.3 & 59.9 & 52.4 & 57.6 & 45.3 & 67.21 \\
H2RBox-v2 \cite{Yu_Yang_Li_Zhou_Zhang_Yan_Da_2023} & \checkmark & 89.0 & 74.4 & 51.0 & 60.5 & 79.8 & 75.3 & 86.9 & 90.9 & 86.1 & 85.0 & 59.2 & 63.2 & 65.2 & 71.6 & 49.7 & \textbf{72.52} \\
\midrule
\multicolumn{17}{@{}l}{\textbf{Point-supervised:}} \\
P2BNet \cite{Chen_Yu_Han_Hassan_Wang_Li_Zhao_Shi_Han_Ye_2022} + H2RBox \cite{Yang_Zhang_Li_Wang_Zhou_Yan_2022} & $\times$ & 24.7 & 35.9 & 7.0 & 27.9 & 3.3 & 12.1 & 17.5 & 17.5 & 0.8 & 34.0 & 6.3 & 49.6 & 11.6 & 27.2 & 18.8 & 19.63 \\
P2BNet \cite{Chen_Yu_Han_Hassan_Wang_Li_Zhao_Shi_Han_Ye_2022} + H2RBox-v2 \cite{Yu_Yang_Li_Zhou_Zhang_Yan_Da_2023} & $\times$ & 11.0 & 44.8 & 14.9 & 15.4 & 36.8 & 16.7 & 27.8 & 12.1 & 1.8 & 31.2 & 3.4 & 50.6 & 12.6 & 36.7 & 12.5 & 21.87 \\
Point2RBox \cite{yu2024point2rbox} & \checkmark & 53.3 & 63.9 & 3.7 & 50.9 & 40.0 & 39.2 & 45.7 & 76.7 & 10.5 & 56.1 & 5.4 & 49.5 & 24.2 & 51.2 & 33.8 & 40.27 \\
PointOBB (FCOSR) \cite{luo2024pointobb} & $\times$ & 26.1 & 65.7 & 9.1 & 59.4 & 65.8 & 34.9 & 29.8 & 0.5 & 2.3 & 16.7 & 0.6 & 49.04 & 21.8 & 41.0 & 36.7 & 30.08 \\
PointOBB (ORCNN) \cite{luo2024pointobb} & $\times$ & 28.3 & 70.7 & 1.5 & 64.9 & 68.8 & 46.8 & 33.9 & 1.0 & 20.1 & 10.0 & 0.2 & 47.0 & 29.7 & 38.2 & 30.6 & 33.31 \\
PointOBB-v2 (FCOSR) \cite{ren2024pointobbv2simplerfasterstronger} & $\times$ & 64.5 & 27.8 & 1.9 & 36.2 & 58.8 & 47.2 & 53.4 & 90.5 & 62.2 & 45.3 & 12.1 & 41.7 & 8.1 & 43.7 & 32.0 & 41.68 \\
PointOBB-v2 (ORCNN) \cite{ren2024pointobbv2simplerfasterstronger} & $\times$ & 63.7 & 45.6 & 2.0 & 39.5 & 50.5 & 49.6 & 45.4 & 89.8 & 62.9 & 41.3 & 13.6 & 42.8 & 8.9 & 39.5 & 29.5 & 41.64\\
\rowcolor{gray!30}
PointOBB-v3 (e2e) & \checkmark & 30.9 & 39.4 & 13.5 & 22.7 & 61.2 & 7.0 & 43.1 & 62.4 & 59.8 & 47.3 & 2.7 & 45.1 & 16.8 & 55.2 & 11.4 & 41.29 \\
\rowcolor{gray!30}
PointOBB-v3 (FCOSR) & $\times$ & 52.9 & 54.4 & 21.3 & 52.7 & 65.6 & 44.9 & 67.8 & 87.2 & 26.7 & 73.4 & 32.6 & 53.3 & 39.0 & 56.4 & 10.2 & 49.24 \\ 
\rowcolor{gray!30}
PointOBB-v3 (ORCNN) & $\times$ & 46.0 & 55.0 & 23.7 & 52.1 & 66.8 & 50.4 & 71.6 & 90.2 & 19.7 & 71.4 & 43.8 & 55.7 & 40.0 & 55.4 & 13.9 & \textbf{50.44} \\
\botrule
\end{tabular*}
\footnotetext{The categories in the DOTA-v1.0 dataset include Plane (PL), Baseball Diamond (BD), Bridge (BR), Ground Track Field (GTF), Small Vehicle (SV), Large Vehicle (LV), Ship (SH), Tennis Court (TC), Basketball Court (BC), Storage Tank (ST), Soccer Ball Field (SBF), Roundabout (RA), Harbor (HA), Swimming Pool (SP), and Helicopter (HC). \textbf{*} Comparison tracks: \checkmark = End-to-end training and testing; $\times$ = Generating pseudo labels to train the detector (two-stage training).}
\end{minipage}
\end{sidewaystable}

\subsection{Main Results}\label{subsec11}
\subsubsection{Results on DIOR-R}\label{subsubsec10}
Tab. \ref{tab:main_dior} compares PointOBB-v3 with existing methods, which can be categorized into two tracks:

\textbf{1) Two-stage mode.} This paradigm generates RBox labels on the training and validation sets and then uses these labels to train oriented object detectors. In the two-stage mode, PointOBB-v3 achieves AP$_{50}$ scores of 40.18\% and 41.82\% when training with Rotated FCOS and Oriented R-CNN, respectively. This marks a notable improvement over Point2RBox (27.34\%), PointOBB (37.31\%/38.08\%) and PointOBB-v2 (34.35\%/39.62\%), setting a more competitive benchmark. 

\textbf{2) End-to-end mode.} This paradigm directly trains a weakly supervised detector on the training and validation sets and performs inference on the test set. Without utilizing any priors, our approach achieves a 10.26\% improvement (37.60\% vs. 27.34\%) over Point2RBox \cite{yu2024point2rbox}. Remarkably, it also demonstrates competitive performance compared to methods in the two-stage mode. For instance, it surpasses PointOBB (FCOSR) by 0.29\% and PointOBB-v2 \cite{ren2024pointobbv2simplerfasterstronger} (FCOSR) by 3.25\%, respectively. Furthermore, as shown in Tab. \ref{tab:epoch}, while maintaining competitive performance, the end-to-end version also achieves improvements in efficiency. Specifically, the number of training epochs is reduced from 36 to 24, and the total training time decreases from 17.23 hours to 13.55 hours—a reduction of approximately 21.36\%. 

\textbf{3) Computational efficiency.} As shown in Tab. \ref{tab:train_cost}, although PointOBB-v3 does have a longer training time compared to other methods, the significant performance improvement compensates for the additional training cost. Since PointOBB-v3 is an extension of PointOBB, it maintains a similar level of efficiency while achieving notable performance improvements. In contrast, extending the training time to 36 epochs on Point2RBox-SK and PointOBB-v2 yields almost no significant performance improvement. Additionally, in terms of inference speed (pseudo-label generation), PointOBB-v3 significantly outperforms PointOBB-v2, especially since PointOBB-v2 requires PCA computation during inference to generate pseudo-labels, which significantly slows down its overall inference speed.

\begin{table}[!tb]
\centering
\caption{Comparison of the training time and the accuracy between the two-stage version and the end-to-end version on the DIOR-R dataset.}
\label{tab:epoch}%
\begin{tabular}{@{}llll@{}}
\toprule
Method & Epochs & Train hours & mAP$_{50}$ \\
\midrule
Two-stage & 36 & 17.23 & \textbf{40.18}\footnotemark[1] \\
End-to-End & \textbf{24} & \textbf{13.55} & 37.60 \\
\botrule
\end{tabular}
\footnotetext[1]{The reported mAP$_{50}$ is trained with Rotated FCOS.}
\end{table}

\begin{figure}[!tb]
    \centering
    \begin{minipage}{0.47\textwidth}
        \centering
        \includegraphics[width=\textwidth]{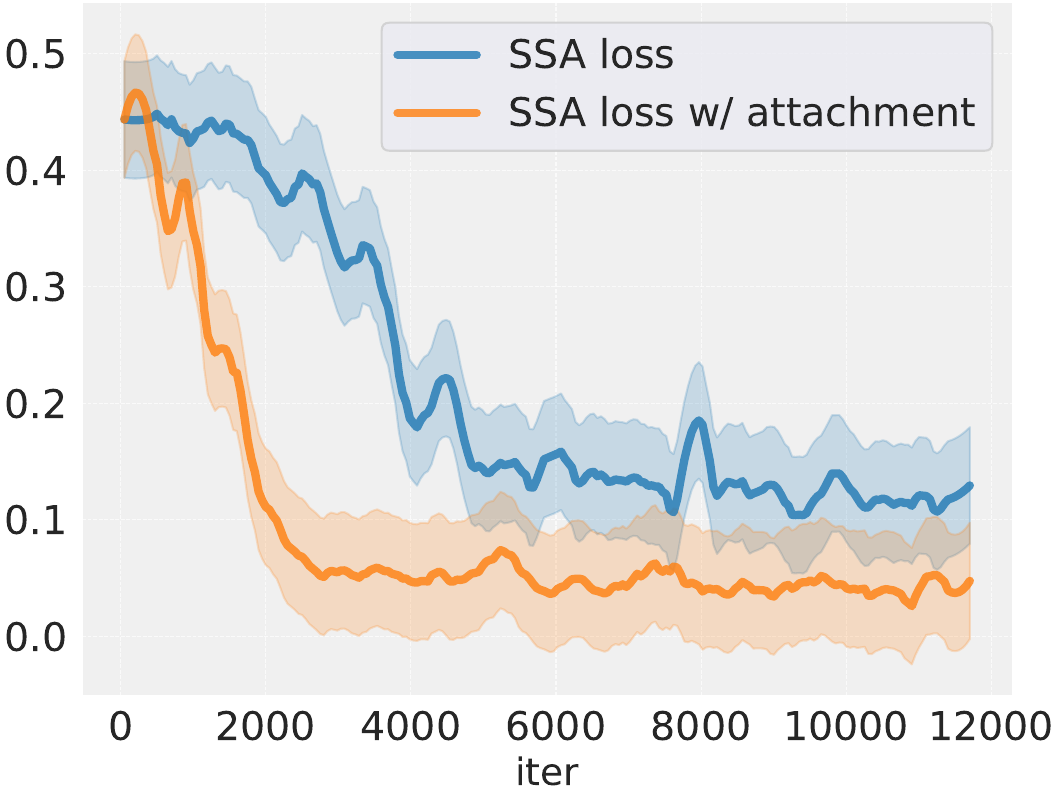}
        \\(a) SSA loss
    \end{minipage}
    \hfill
    \begin{minipage}{0.47\textwidth}
        \centering
        \includegraphics[width=\textwidth]{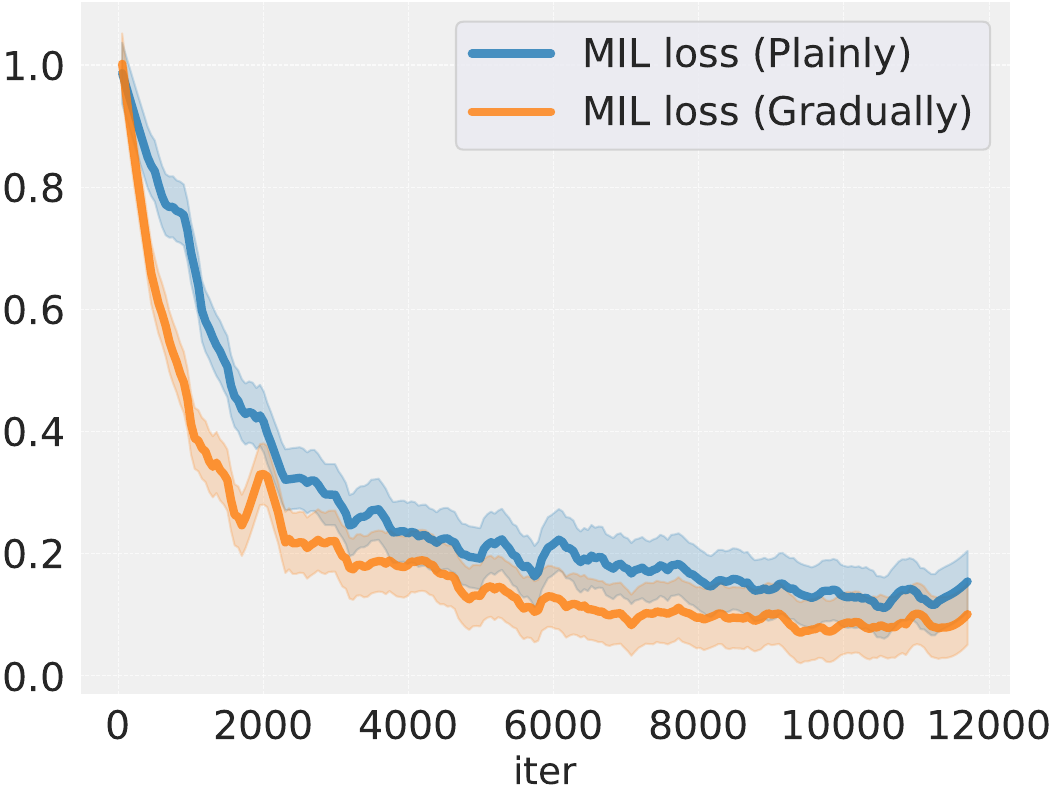}
        \\(b) MIL loss
    \end{minipage}
    \caption{(a) indicates the SSA loss under different settings. In (b), ``MIL loss (Gradually)" means training the network with the proposed gradual strategy (i.e., in chronological order), and ``MIL loss (Plainly)" means training the angle branch from the beginning directly.}
    \label{fig:LossCurve}
\end{figure}

\subsubsection{Results on DOTA-v1.0}\label{subsubsec11}
As shown in Tab. \ref{tab:main_dota10}, the comparison of PointOBB-v3 with existing methods is organized into two distinct tracks:

\textbf{1) Two-stage mode.} PointOBB-v3 achieves an improvement compared to existing point-supervised approaches. The two-stage version of PointOBB-v3 achieves AP$_{50}$ of 49.24\% and 50.44\% by training Rotated FCOS and Oriented R-CNN on DOTA-v1.0. Specifically, PointOBB-v3 achieves an improvement over PointOBB, surpassing it by 17.13\%, and demonstrates an 8.80\% advantage over PointOBB-v2 when training under the Oriented R-CNN. 

\textbf{2) End-to-end mode.} The end-to-end version of PointOBB-v3 attains AP$_{50}$ scores of 41.29\% on DOTA-v1.0. Specifically, it achieves an improvement of 11.21\% over PointOBB (FCOSR) and maintains comparable performance to PointOBB-v2 across its two detectors. Additionally, when compared to Point2RBox, which leverages manual sketches to assist in boundary determination, our approach achieves a remarkable gain of 1.02\%.

Additionally, the visual result is shown in Fig.~\ref{fig:show}. The proposed PointOBB-v3 effectively addresses the issue of local focus MIL fashion and accurately predicts the object orientation.

\begin{sidewaystable}
\centering
\captionsetup{justification=raggedright, singlelinecheck=false}
\caption{Results on more datasets.}
\label{tab:else_results}
\begin{tabular}{l|c|c|c|c|c|c|c}
\hline
\textbf{Method} & \textbf{DIOR} & \textbf{DOTA-v1.0} & \textbf{DOTA-v1.5} & \textbf{DOTA-v2.0} & \textbf{FAIR1M} & \textbf{STAR} & \textbf{RSAR} \\ \hline
Point2Mask-RBox \cite{Li_Yuan_Wang_Zhu_Li_Liu_Zhang} & 13.77 & 9.72 & - & - & - & - & - \\ 
P2BNet \cite{Chen_Yu_Han_Hassan_Wang_Li_Zhao_Shi_Han_Ye_2022} + H2RBox \cite{Yang_Zhang_Li_Wang_Zhou_Yan_2022} & 22.59 & 19.63 & - & - & - & - & - \\ 
P2BNet \cite{Chen_Yu_Han_Hassan_Wang_Li_Zhao_Shi_Han_Ye_2022} + H2RBox-v2 \cite{Yu_Yang_Li_Zhou_Zhang_Yan_Da_2023} & 23.61 & 21.87 & - & - & - & - & - \\ 
Point2RBox-RC \cite{yu2024point2rbox} & 24.66 & 34.07 & 24.31 & 14.69 & 8.08 & 7.86 & - \\ 
Point2RBox-SK* \cite{yu2024point2rbox} & 27.34 & 40.27 & 30.51 & \textbf{23.43} & \textbf{20.03} & - & - \\
\rowcolor{gray!20} PointOBB-v3 (e2e) & \textbf{37.60} & \textbf{41.29} & \textbf{31.25} & 22.82 & 11.42 & \textbf{11.31} & \textbf{15.84} \\ \hline
PointOBB (FCOSR) \cite{luo2024pointobb} & 37.31 & 30.08 & 10.66 & 5.53 & 11.19 & 9.19 & 13.80 \\ 
PointOBB-v2 (FCOSR) \cite{ren2024pointobbv2simplerfasterstronger} & 34.35 \textcolor[rgb]{0.6,0.6,0.6}{\tiny (-2.96)} & 41.68 \textcolor[rgb]{0,0.6,0}{\tiny (+11.60)} & 30.59 \textcolor[rgb]{0,0.6,0}{\tiny (+19.93)} & 20.64 \textcolor[rgb]{0,0.6,0}{\tiny (+15.11)} & 13.36 \textcolor[rgb]{0,0.6,0}{\tiny (+2.17)} & 9.00 \textcolor[rgb]{0.6,0.6,0.6}{\tiny (-0.19)} & 18.99 \textcolor[rgb]{0,0.6,0}{\tiny (+5.19)} \\
\rowcolor{gray!20} PointOBB-v3 (FCOSR) & \textbf{40.18} \textcolor[rgb]{0,0.6,0}{\tiny (+2.87)} & \textbf{49.24} \textcolor[rgb]{0,0.6,0}{\tiny (+19.16)} & \textbf{33.79} \textcolor[rgb]{0,0.6,0}{\tiny (+23.13)} & \textbf{23.52} \textcolor[rgb]{0,0.6,0}{\tiny (+17.99)} & \textbf{18.35} \textcolor[rgb]{0,0.6,0}{\tiny (+7.16)} & \textbf{12.85} \textcolor[rgb]{0,0.6,0}{\tiny (+3.66)} & \textbf{22.60} \textcolor[rgb]{0,0.6,0}{\tiny (+8.80)} \\ 
\hline
PointOBB (ORCNN) \cite{luo2024pointobb} & 38.08 & 33.31 & 10.92 & 6.29 & 12.02 & 10.66 & 12.30 \\ 
PointOBB-v2 (ORCNN) \cite{ren2024pointobbv2simplerfasterstronger} & 39.62 \textcolor[rgb]{0,0.6,0}{\tiny (+1.54)} & 41.64 \textcolor[rgb]{0,0.6,0}{\tiny (+8.33)} & 32.01 \textcolor[rgb]{0,0.6,0}{\tiny (+21.09)} & 23.40 \textcolor[rgb]{0,0.6,0}{\tiny (+17.11)} & 13.42 \textcolor[rgb]{0,0.6,0}{\tiny (+1.40)} & 8.14 \textcolor[rgb]{0.6,0.6,0.6}{\tiny (-2.52)} & 22.61 \textcolor[rgb]{0,0.6,0}{\tiny (+10.31)} \\ 
\rowcolor{gray!20} PointOBB-v3 (ORCNN) & \textbf{41.82} \textcolor[rgb]{0,0.6,0}{\tiny (+3.74)} & \textbf{50.44} \textcolor[rgb]{0,0.6,0}{\tiny (+17.13)} & \textbf{38.08} \textcolor[rgb]{0,0.6,0}{\tiny (+27.16)} & \textbf{24.86} \textcolor[rgb]{0,0.6,0}{\tiny (+18.57)} & \textbf{20.19} \textcolor[rgb]{0,0.6,0}{\tiny (+8.17)} & \textbf{16.73} \textcolor[rgb]{0,0.6,0}{\tiny (+6.07)} & \textbf{22.84}
\textcolor[rgb]{0,0.6,0}{\tiny (+10.54)} \\
\hline
\end{tabular}
\footnotetext{Results on DIOR, DOTA-v1.0/v1.5/v2.0, FAIR1M, STAR and RSAR datasets, reporting the mAP$_{50}$ metric. The term e2e refers to the end-to-end version of the framework. Rotated FCOS and Oriented R-CNN refer to the two-stage version of the framework, using Rotated FCOS and Oriented R-CNN as detectors, respectively. * indicates using additional human knowledge priors.}
\end{sidewaystable}

\begin{table}[!tb]
\centering
\caption{The effects of SSC loss, DS matching strategy and SSFF in DIOR-R and DOTA-v1.0 datasets. We evaluate Rotated FCOS on the DIOR-R testing set and Oriented R-CNN on the DOTA-v1.0 testing set.}
\label{tab:SSC&DS&SSFF}
\begin{tabular}{ccc|cc|cc}
\toprule
\multicolumn{3}{c|}{\textbf{Module}} & \multicolumn{2}{c|}{\textbf{DIOR-R} \cite{Cheng_2022}} & \multicolumn{2}{c}{\textbf{DOTA-v1.0} \cite{Xia_Bai_Ding_Zhu_Belongie_Luo_Datcu_Pelillo_Zhang_2018}} \\ \midrule
\textbf{SSC} & \textbf{DS} & \textbf{SSFF} & \textbf{mIoU} & \textbf{mAP$_{50}$} & \textbf{mIoU} & \textbf{mAP$_{50}$} \\ \midrule
 &  &  & 47.95 & 30.16 & 40.26 & 28.44 \\ 
 & \checkmark &  & 50.78 & 31.96 & 42.54 & 30.63 \\
\checkmark &  &  & 53.17 & 36.39 & 43.92 & 32.98 \\ 
\checkmark & \checkmark &  & 56.08 & 37.31 & 45.35 & 33.31 \\
\rowcolor{gray!20} \checkmark & \checkmark & \checkmark & \textbf{57.68} & \textbf{40.18} & \textbf{51.33} & \textbf{50.44} \\
\bottomrule
\end{tabular}
\vspace{-5pt}
\end{table}

\begin{table}[!tb]
\centering
\begin{minipage}[t]{0.49\linewidth}
\setlength{\tabcolsep}{9pt} 
\centering
\caption{Ablation studies of using MIL loss at different views on DIOR-R. mAP is reported by training the Oriented R-CNN. ``O'' indicates the original view, ``R'' indicates the resized view, ``R/F'' indicates the Rot/Flp view.}
\label{tab:mil_loss_views}
\begin{tabular}{ccccc}
\hline
\textbf{O} & \textbf{R} & \textbf{R/F} & \textbf{mIoU} & \textbf{mAP}$_{50}$ \\ \hline
\checkmark &           &           & 52.90 & 34.94 \\ 
\checkmark & \checkmark &           & 54.44 & 36.26 \\ 
\checkmark &           & \checkmark & 53.49 & 35.80 \\ 
\rowcolor{gray!20} \checkmark & \checkmark & \checkmark & \textbf{56.08} & \textbf{38.08} \\ \hline
\end{tabular}
\end{minipage}
\hfill
\begin{minipage}[t]{0.48\linewidth}
\setlength{\tabcolsep}{10pt} 
\centering
\caption{Ablation studies of the burn-in steps on DIOR-R. mAP is reported by training the Oriented R-CNN. ``P'': Plainly, ``G'': Gradually, ``A'' indicates attaching gradients between the angle branch and the other parts.}
\label{tab:burn_in_steps}
\begin{tabular}{ccccc}
\hline
\textbf{P} & \textbf{G} & \textbf{A} & \textbf{mIoU} & \textbf{mAP}$_{50}$ \\ \hline
\checkmark &           &           & 34.92 & 23.71 \\ 
\checkmark &           & \checkmark & 37.45 & 25.52 \\ 
          & \checkmark &           & 48.92 & 32.23 \\ 
\rowcolor{gray!20}     & \checkmark & \checkmark & \textbf{56.08} & \textbf{38.08} \\ \hline
\end{tabular}
\end{minipage}
\end{table}

\subsection{Results on More Datasets}\label{subsubsec12}
The results are displayed in Tab. \ref{tab:else_results}. On the more challenging DOTA-v1.5 dataset, PointOBB-v3 exhibits a comparable trend, outperforming PointOBB by 23.13\%/27.16\% and surpassing PointOBB-v2 by 3.2\%/6.07\% in the pseudo-generation track. Similarly, on DOTA-v2.0, PointOBB-v3 achieves a 17.99\%/18.57\% improvement over PointOBB and a 2.88\%/1.46\% gain over PointOBB-v2. Our method also demonstrates strong performance on various datasets such as FAIR1M, STAR and RSAR. Specifically, PointOBB-v3 achieves improvements of 7.16\%/3.66\%/8.80\% compared to PointOBB and 4.99\%/3.85\%/3.61\% over PointOBB-v2 with the Rotated FCOS detector.

On the other hand, the end-to-end version of PointOBB-v3 also delivers strong performance on these more challenging datasets. Specifically, it outperforms Point2RBox-RC on DOTA-v1.5, DOTA-v2.0, FAIR1M, and STAR datasets. Even when compared to Point2RBox-SK, which incorporates human prior knowledge, our method also demonstrates highly competitive performance, particularly achieving a certain degree of superiority on the DIOR, DOTA-v1.0, and DOTA-v1.5 datasets (37.60\% vs. 27.34\%, 41.29\% vs. 40.27\%, 31.25\% vs. 30.51\%). Moreover, on the STAR dataset, our end-to-end version outperforms PointOBB (FCOSR) and PointOBB-v2 (FCOSR) by 2.12\% and 2.31\%, respectively, while achieving gains of 2.04\% and 4.28\% on the RSAR dataset. It is evident that the end-to-end version still has a certain performance gap compared to the two-stage version. The potential reasons for this gap are as follows: (1) In the end-to-end version, the detection branch and the MIL branch are jointly trained. However, the training objectives of these two branches differ, which can lead to conflicting gradients during optimization, ultimately affecting overall performance. (2) Additionally, in the end-to-end version, the supervision signal for the detection branch is less stable compared to the two-stage version. This instability in the supervision signal can cause less consistent convergence during training, making it more challenging for the network to reach optimal performance.

\subsection{Ablation Studies}\label{subsec12}

\subsubsection{Two-Stage Framework}\label{subsubsec13}
\noindent\textbf{The effect of SSC, SSFF, and DS.}
Tab. \ref{tab:SSC&DS&SSFF} examines the impact of the three proposed key components: the SSC loss, the SSFF module, and the DS matching strategy. As illustrated in Tab. \ref{tab:SSC&DS&SSFF}, each of these elements plays a role in enhancing performance. This demonstrates that the scale-consistency-based SSC loss and the scale-aware SSFF module have a substantial impact, while the DS matching strategy further improves performance by addressing misalignment caused by unknown scales.

\noindent\textbf{The effect of different MIL losses.} Tab. \ref{tab:mil_loss_views} highlights the impact of applying MIL loss across various views. The results indicate that both enhanced views contribute to accuracy improvements, likely due to their representation of distinct types of data augmentation.

\noindent\textbf{The effect of the angle branch's setting.} We investigate the effect of gradient backpropagation on the angle branch within the angle acquisition module. As presented in Fig. \ref{fig:LossCurve}(a) and Tab. \ref{tab:burn_in_steps}, optimizing the angle alongside the base network (``Attachment") results in quicker convergence and improved accuracy compared to the strategy without ``Attachment’', highlighting the importance of collaborative optimization between angle and scale.

\noindent\textbf{The effect of the burn-in steps.} Building on the ``Attachment" setting, we examine the influence of burn-in steps through MIL loss. In the ``Plainly" strategy, burn-in steps are set to 0, meaning angle learning is introduced from the start. Conversely, the ``Gradually" strategy applies the defined burn-in steps progressively. As indicated in Fig. \ref{fig:LossCurve}(b) and Tab. \ref{tab:burn_in_steps}, the ``Plainly" strategy affects the initial optimization of the MIL network, resulting in performance degradation. In contrast, the ``Gradually" strategy incorporates two burn-in steps after scale and angle learning have mostly converged, enabling the network to optimize progressively and achieve better results. Furthermore, as shown in Tab. \ref{tab:burn_in_steps2}, we experimented with multiple burn-in step settings and observed that the model's performance is significantly affected by these variations. Notably, the settings of ``6'' and ``8'' proved to be more effective in facilitating the optimization of the model.  And the settings of ``4'', ``8'' and ``8'', ``10'' may have led to suboptimal performance of the network due to insufficient scale or angle learning.

\noindent\textbf{The effect of point range.}
The right section of Tab. \ref{tab:grouping_type&point_range} demonstrates the effect of varying ranges during point label generation. Adding appropriate noise shows benefits over using only the center point label (i.e., 0\%). This is likely because, for certain categories, relying solely on the center point limits the network's perceptual range, reducing its ability to recognize object boundaries effectively.

\noindent\textbf{The effect of grouping type in SSC loss.} The left section of Tab. \ref{tab:grouping_type&point_range} evaluates the influence of grouping scores by scale, ratio, and proposal within the SSC loss. The results confirm the effectiveness of scale-based grouping, supporting the initial objective of our design.

\begin{table}[!tb]
\centering
\renewcommand{\arraystretch}{1.3}
\caption{Ablation studies of the grouping type used in the SSC loss, and the generation range of point annotation on DIOR-R. mAP is reported by training the Oriented R-CNN.}
\label{tab:grouping_type&point_range}
\begin{tabular}{c|ccc|ccc}
\hline
{\textbf{Parameter}} & \multicolumn{3}{c|}{\textbf{Grouping Type}} & \multicolumn{3}{c}{\textbf{Point Range}} \\ \hline
\textbf{Setting} & proposal & ratio & \cellcolor{gray!20} scale & 0\% & \cellcolor{gray!20} 10\% & 20\% \\ \hline
\textbf{mIoU}   & 43.95 & 54.38 & \cellcolor{gray!20} \textbf{56.08} & 49.47 & \cellcolor{gray!20} \textbf{56.08} & 54.21 \\ 
\textbf{mAP$_{50}$}  & 30.42 & 36.71 & \cellcolor{gray!20} \textbf{38.08} & 32.11 & \cellcolor{gray!20} \textbf{38.08} & 35.78 \\ \hline
\end{tabular}
\end{table}

\begin{table}[!tb]
\centering
\begin{minipage}[t]{0.49\linewidth}
\setlength{\tabcolsep}{10pt} 
\centering
\caption{Ablation studies of the burn-in steps. The terms step1 and step2 refer to burn-in step1 and burn-in step2 in the progressive multi-view switching strategy, respectively. The results are evaluated on the DIOR-R dataset. mAP is reported by training the Oriented R-CNN.}
\label{tab:burn_in_steps2}
\begin{tabular}{cccc}
\toprule
\textbf{step1} & \textbf{step2} & \textbf{mIoU} & \textbf{mAP$_{50}$} \\
\midrule
 0 & 0 & 53.53 & 38.33 \\
 0 & 8 & 55.59 & 39.59 \\
\rowcolor{gray!20} 6 & 8 & \textbf{57.68} & \textbf{41.82} \\
 4 & 8 & 54.69 & 39.42 \\
 8 & 10 & 53.91 & 38.22 \\
\bottomrule
\end{tabular}
\end{minipage}
\hfill
\begin{minipage}[t]{0.48\linewidth}
\setlength{\tabcolsep}{6pt} 
\centering
\caption{The effects of SSFF, and IAW in DIOR-R and DOTA-v1.0 datasets. We evaluate the end-to-end detector on the DIOR-R testing set and the DOTA-v1.0 testing set.}
\label{tab:e2e}
\renewcommand{\arraystretch}{1.3} 
\begin{tabular}{cc|c|c}
\toprule
\multicolumn{2}{c|}{\textbf{Module}} & \textbf{DIOR-R} & \textbf{DOTA-v1.0} \\
\midrule
\textbf{SSFF} & \textbf{IAW} & \textbf{mAP$_{50}$} & \textbf{mAP$_{50}$} \\
\midrule
 &  & 25.76  & 24.91 \\ 
 & \checkmark & 27.79 & 27.37  \\
\checkmark &  & 34.88 & 35.74  \\ 
\rowcolor{gray!20} \checkmark & \checkmark & \textbf{37.60} & \textbf{41.29} \\
\bottomrule
\end{tabular}
\end{minipage}
\end{table}

\subsubsection{End-to-End Framework}\label{subsubsec14}
\noindent\textbf{The effect of SSFF and IAW.} As shown in Tab. \ref{tab:e2e}, the proposed SSFF and IAW also play a significant role in the end-to-end version of PointOBB-v3. Compared to not using them, our model's performance improves by 11.84\% on the DIOR dataset and 16.38\% on the DOTA-v1.0 dataset. This demonstrates that SSFF and IAW are indispensable components of our proposed method.

\begin{table*}[t]
\caption{Comparison of computational efficiency on the DIOR-R dataset, including training epochs, training hours, number of parameters, and inference speed. ``FPS” indicates the inference speed of the trained model in generating pseudo-labels. For the two-stage methods, the values in ``()" and the mAP represent the detector's performance (FCOSR).}\label{tab:train_cost}
\fontsize{9pt}{12pt}\selectfont
\setlength{\tabcolsep}{1.1mm}
\setlength{\aboverulesep}{0.4ex}
\setlength{\belowrulesep}{0.4ex}
\setlength{\abovecaptionskip}{1.5mm}
\centering
\begin{tabular}{l|ccccc}
\toprule
\textbf{Method} & \textbf{Epochs} & \textbf{Train hours} & \textbf{Params} & \textbf{FPS} & \textbf{mAP\textsubscript{50}} \\
\hline
\rowcolor{gray!20} \multicolumn{6}{l}{$\blacktriangledown$ \textbf{End-to-end:}} \\ \hline
Point2RBox-SK & 12 & 3.78 & 52.92M & 33.6 & 27.34 \\
Point2RBox-SK & 24 & 7.56 & 52.92M & 33.6 & 28.50 \\
Point2RBox-SK & 36 & 11.34 & 52.92M & 33.6 & 28.90 \\
PointOBB-v3 (e2e) & 24 & 13.55 & 46.58M & 33.6 & 37.60 \\
\hline
\rowcolor{gray!20} \multicolumn{6}{l}{$\blacktriangledown$ \textbf{Two-stage:}} \\ \hline
PointOBB & 24 (12) & 11.70 (2.48) & 41.80M (31.93M) & 26.1 (33.6) & 37.31 \\
PointOBB-v2 & 12 (12) & 2.67 (2.48) & 32.59M (31.93M) & 5.9 (33.6) & 34.35 \\
PointOBB-v2 & 24 (12) & 5.34 (2.48) & 32.59M (31.93M) & 5.9 (33.6) & 25.88 \\
PointOBB-v3 (two stage) & 24 (12) & 14.75 (2.48) & 41.80M (31.93M) & 23.7 (33.6) & 40.18 \\
\hline
\end{tabular}
\end{table*}

\begin{figure}[h]
    \centering
    \includegraphics[width=1\textwidth]{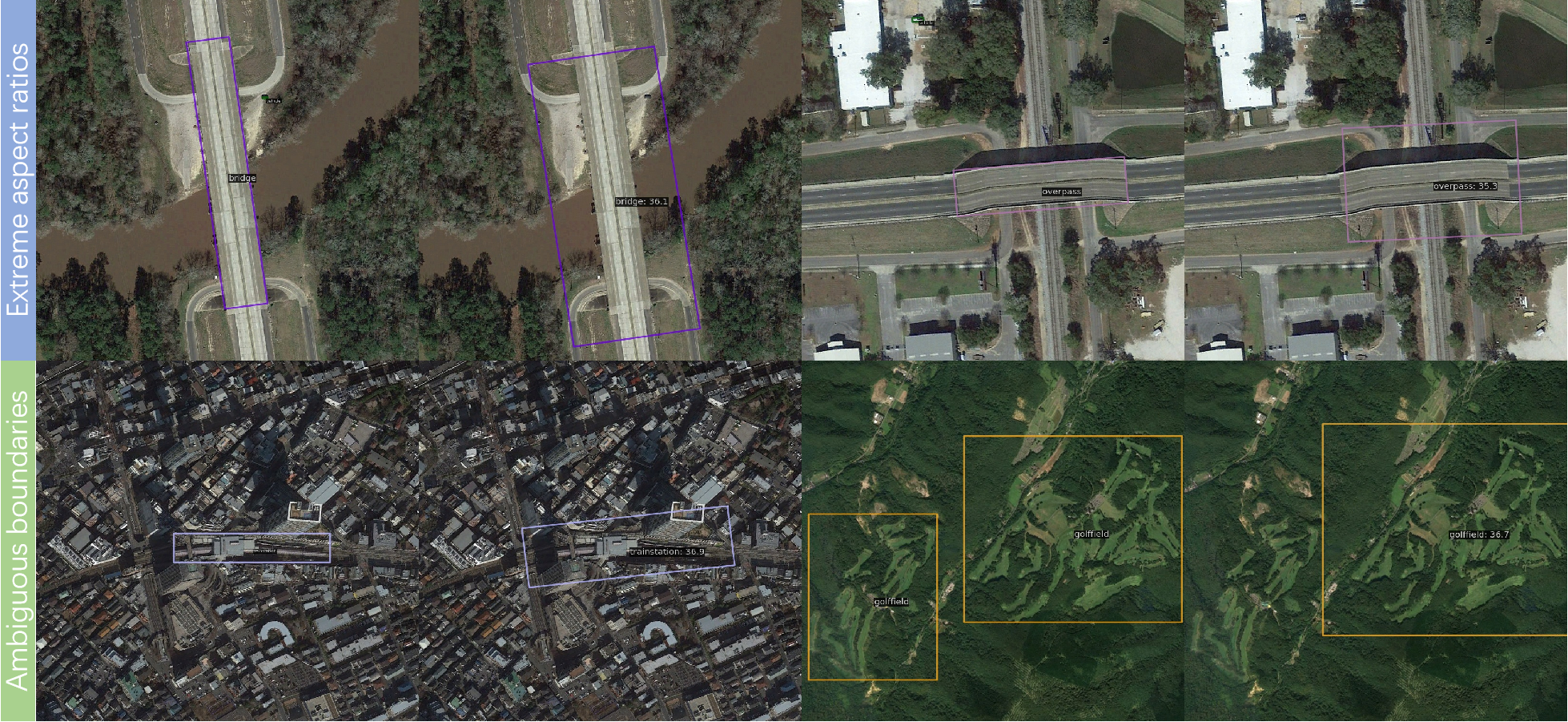}
    \caption{Qualitative analysis on struggling cases. On the left side of each image is the ground truth, and on the right side is the prediction result.}
    \label{fig:bad_case}
\end{figure}

\section{Conclusions}\label{sec6}
This paper introduces PointOBB-v3, a stronger point-based OBB generation framework for OOD. PointOBB-v3 effectively captures the scale and orientation of objects by leveraging three unique views and employing a progressive multi-view switching strategy. With these views, we design a scale augmentation module and an angle acquisition module. The scale augmentation module improves the network's ability to recognize object scales by integrating a Scale-Sensitive Consistency (SSC) loss and a Scale-Sensitive Feature Fusion (SSFF) module. The angle acquisition module enables self-supervised angle learning and enhances the precision of object angle prediction through Dense-to-Sparse (DS) matching. Additionally, for the end-to-end version, we propose the Instance-Aware Weighting (IAW) strategy to optimize the joint training of the two branches. Our method outperforms current approaches on the DIOR-R, DOTA-v1.0/v1.5/v2.0, FAIR1M, STAR, and RSAR datasets, achieving an average improvement of 3.56\% across all datasets. We hope this work could foster further progress in the area of single point-supervised OOD.

\noindent\textbf{Future work.} There still remains a gap between PointOBB-v3 and RBox-supervised OOD, particularly for specific categories such as BR and OP, which are challenged by extreme aspect ratios. (as shown in Fig. \ref{fig:bad_case}) The long, narrow shapes of these objects make them highly sensitive to small changes in angle. In addition, categories like GF and TS struggle due to unclear or ambiguous object boundaries, making accurate detection difficult. This highlights the need for further investigation into the use of spatial and contextual features in aerial imagery to address these challenges.

\section*{Data Availability}
The datasets used in this paper are publicly available. Their names and links are as follows:

1. DIOR \href{https://gcheng-nwpu.github.io/}{https://gcheng-nwpu.github.io/}

2. DOTA-V1.0/V1.5/V2.0 \href{https://captain-whu.github.io/DOTA/index.html}{https://captain-whu.github.io/DOTA/index.html}

3. FAIR1M \href{https://www.gaofen-challenge.com/benchmark}{https://www.gaofen-challenge.com/benchmark}

4. STAR \href{https://linlin-dev.github.io/project/STAR}{https://linlin-dev.github.io/project/STAR}

5. RSAR \href{https://github.com/zhasion/RSAR}{https://github.com/zhasion/RSAR}

\section*{Acknowledgments}
This work was supported by the National Natural Science Foundation of China under Grants (42371321; 42030102) and the Ant Group.

\bibliography{sn-bibliography}

\end{document}